%% file: main.tex
\pdfoutput=1

\documentclass[11pt]{article}

\usepackage[]{ACL2023}

\usepackage{times}
\usepackage{latexsym}

\usepackage{hyperref}
\usepackage{url}
\usepackage{multirow}
\usepackage{makecell}
\usepackage{subfigure}
\usepackage{color}
\usepackage{booktabs}
\usepackage{adjustbox}
\usepackage{color}
\usepackage{xspace}
\usepackage{todonotes}
\usepackage{paralist}
\usepackage{amsmath}
\usepackage{balance} 
\usepackage{multicol}
\usepackage{amsfonts} 

\newtheorem{finding}{Observation}

\newcommand{\fb}{\texttt{FB15K}\xspace}
\newcommand{\wn}{\texttt{WN18}\xspace}
\newcommand{\wnrr}{\texttt{WN18RR}\xspace}
\newcommand{\fbt}{\texttt{FB15K-237}\xspace}
\newcommand{\nell}{\texttt{NELL-995}\xspace}

\usepackage[T1]{fontenc}

\usepackage[utf8]{inputenc}

\usepackage{microtype}

\usepackage{inconsolata}

\title{Are Message Passing Neural Networks Really Helpful for Knowledge Graph Completion?}

\author{Juanhui Li$^1$,  Harry Shomer$^1$ ,  Jiayuan Ding$^1$ , Yiqi Wang$^1$\thanks{Corresponding Author},  Yao Ma$^2$ 
\\ {\bf Neil Shah$^3$},  {\bf Jiliang Tang$^1$},
 {\bf Dawei Yin$^4$}\\ 
$^{1}$Michigan State University, 
$^{2}$New Jersey Institute of Technology \\
$^{3}$Snap Inc., $^{4}$Baidu Inc.\\
\texttt{\{lijuanh1,shomerha,dingjia5,wangy206,tangjili\}@msu.edu }\\
\texttt{yao.ma@njit.edu, nshah@snap.com, yindawei@acm.org }
}

\begin{document}
\maketitle
\begin{abstract}Knowledge graphs (KGs) facilitate a wide variety of applications. Despite great efforts in creation and maintenance, even the largest KGs are far from complete. Hence, KG completion (KGC) has become one of the most crucial tasks for KG research. Recently, considerable literature in this space has centered around the use of Message Passing (Graph) Neural Networks (MPNNs), to learn powerful embeddings.  The success of these methods is naturally attributed to the use of MPNNs over simpler multi-layer perceptron (MLP) models, given their additional message passing (MP) component. In this work, we find that surprisingly, simple MLP models are able to achieve comparable performance to MPNNs, suggesting that MP may not be as crucial as previously believed. With further exploration, we show careful scoring function and loss function design has a much stronger influence on KGC model performance. This suggests a conflation of scoring function design, loss function design, and MP in prior work, with promising insights regarding the scalability of state-of-the-art KGC methods today, as well as careful attention to more suitable MP designs for KGC tasks tomorrow. Our codes are publicly available at: \url{https://github.com/Juanhui28/Are_MPNNs_helpful}.
\end{abstract}

\input{1_introduction}
\input{3_preliminaries}
\input{4_primary_findings}

\input{5_discussion}

\input{2_related_work}

\input{6_conclusion}
\section{Limitations}

In this paper, we conducted investigation on MPNN-based KGC models. 
MPNN-based models learn the node representations through aggregating from the local neighborhood, which differ from some recent path-based works that learn pair-wise representations by integrating the path information between the node pair. Moreover, we mainly focus on the KGC task which is based on knowledge graph, and thus other types of graph (e.g., homogeneous graph) are not considered. Therefore, our findings and observations might not be applicable for other non-MPNN-based models and non-KGC task.




\bibliography{anthology,custom}
\bibliographystyle{acl_natbib}

\appendix

\input{appendix}


\end{document}

%% file: 1_introduction.tex
\section{Introduction}

Knowledge graphs (KGs) \citep{bollacker2008freebase,carlson2010toward} are a type of knowledge base, which store multi-relational factual knowledge in the form of triplets. Each triplet specifies the relation between a head and a tail entity. KGs conveniently capture rich structured knowledge about many types of entities (e.g. objects, events, concepts) and thus facilitate numerous applications such as information retrieval~\citep{xiong2017explicit}, recommender systems~\citep{wang2019kgat}, and question answering \citep{west2014knowledge}. To this end, the adopted KGs are expected to be as comprehensive as possible to provide all kinds of required knowledge. However, existing large-scale KGs are known to be far from complete with large portions of triplets missing~\citep{bollacker2008freebase,carlson2010toward}. Imputing these missing triplets is of great importance. Furthermore, new knowledge (triplets) is constantly emerging even between existing entities, which also calls for dedicated efforts to predict these new triplets~\citep{garcia2018learning,jin2019recurrent}. Therefore, \emph{knowledge graph completion} (KGC) is a problem of paramount importance~\citep{lin2015learning,yu2021knowledge}. A crucial step towards better KGC performance is to learn low-dimensional continuous embeddings for both entities and relations~\citep{BordesUGWY13}.


Recently, due to the intrinsic graph-structure of KGs, Graph Neural Networks (GNNs) have been adopted to learning more powerful embeddings for their entities and relations, and thus facilitate the KGC. There are mainly two types of GNN-based KGC methods: Message Passing Neural Networks (MPNNs)~\citep{schlichtkrull2018modeling,VashishthSNT20} and path-based methods~\citep{zhu2021neural,zhang2022knowledge,zhu2022learning}. In this work, we focus on MPNN-based models, which
 update node features through a message passing (MP) process over the graph where each node collects and transforms features from its neighbors. When adopting MPNNs for KGs, dedicated efforts are often devoted to developing more sophisticated MP processes that are customized for better capturing multi-relational information~\citep{VashishthSNT20,schlichtkrull2018modeling,ye2019vectorized}. The improvement brought by MPNN-based models is thus naturally attributed to these enhanced MP processes.
Therefore, current research on developing better MPNNs for KGs is still largely focused on advancing MP processes. 


\textbf{Present Work.} In this work, we find that, surprisingly, the MP in the MPNN-based models is not the most critical reason for  reported performance improvements for KGC. 
Specifically, we replaced MP in several state-of-the-art KGC-focused MPNN models such as RGCN~\citep{schlichtkrull2018modeling},  CompGCN~\citep{VashishthSNT20} and KBGAT~\citep{nathani2019learning} with simple Multiple Layer Perceptrons (MLPs) and achieved comparable performance to their corresponding MPNN-based models, across a variety of datasets and implementations. We carefully scrutinized  these MPNN-based models and discovered they also differ from each other in other key components such as scoring functions and loss functions. To better study how these components contribute to the model, we conducted comprehensive experiments to demonstrate the effectiveness of each component. Our results indicate that the scoring and loss functions have stronger influence while MP makes almost no contributions. Based on our findings, we develop ensemble models built upon MLPs, which are able to achieve better performance than MPNN-based models; these implications are powerful in practice, given scalability advantages of MLPs over MPNNs~\citep{zhang2021graphless}.

%% file: 3_preliminaries.tex
\section{Preliminaries}
Before moving to main content, we first introduce KGC-related preliminaries, five datasets and three MPNN-based models we adopt for investigations.

\subsection{Knowledge graph completion (KGC)}
The task of KGC is to infer missing triplets based on known facts in the KG. In KGC, we aim to predict a missing head or tail entity given a triplet. Specifically, we denote the triplets with missing head (tail) entity as $(h,r,?)$ ($(?,r,t)$), where the question mark indicates the entity we aim to predict. Since the head entity prediction and tail entity prediction tasks are symmetric, in the following, we only use the tail entity prediction task for illustration. When conducting the KGC task for a triplet $(h,r,?)$, we use all entities in KG as candidates and try to select the best one as the tail entity. Typically, for each candidate entity $t'$, we evaluate its score for the triplet $(h,r,t')$ with the function $s_{h,r}(t') = f(h,r,t')$, 
where $s_{h,r}(t')$ is the score of $t'$ given the head entity $h$ and the relation $r$, and $f$ is a scoring function. We choose the entity $t'$ with the largest score as the predicted tail entity. $f(\cdot)$ can be modeled in various ways as discussed later. 

\noindent\textbf{Datasets.}
We use five well-known KG datasets, i.e., \textbf{FB15k}~\cite{BordesUGWY13}, \textbf{FB15k-237} \cite{toutanova2015representing,toutanova2015observed}, \textbf{WN18} \cite{schlichtkrull2018modeling}, \textbf{WN18RR} \cite{ettmersMS018} and \textbf{NELL-995}~\citep{XiongHW17} for this study. The detailed descriptions and data statistics can be found in {\bf Appendix~\ref{sec:data}}. Following the settings in previous works~\citep{VashishthSNT20,schlichtkrull2018modeling}, triplets in these datasets are randomly split into training, validation, and test sets, denoted $\mathcal{D}_{train},\mathcal{D}_{val}, \mathcal{D}_{test}$, respectively. The triplets in the training set are regarded as the known facts. We manually remove the head/tail entity of the triplets in the validation and test sets for model selection and evaluation. Specifically, for the tail entity prediction task, given a triplet $(h,r,t^*)$, we remove $t^*$ and construct a test sample $(h,r,?)$. The tail entity $t^*$ is regarded as the ground-truth for this sample. 



\noindent\textbf{Evaluation Metrics.}
When evaluating the performance, we focus on the predicted scores for the ground-truth entity of the triplets in the test set $\mathcal{D}_{test}$. For each triplet $(h,r,?)$ in the test set, we sort all candidate entities $t$ in a non-increasing order according to $s_{h,r}(t)$. Then, we use the rank-based measures to evaluate the prediction quality, including Mean Reciprocal Rank (\textbf{MRR}) and \textbf{Hits@N}. In this work, we choose $N\in\{1,3,10\}$. See {\bf Appendix~\ref{sec:metric}} for their definitions.




\begin{figure}
  \centering
  \includegraphics[width=1\linewidth]{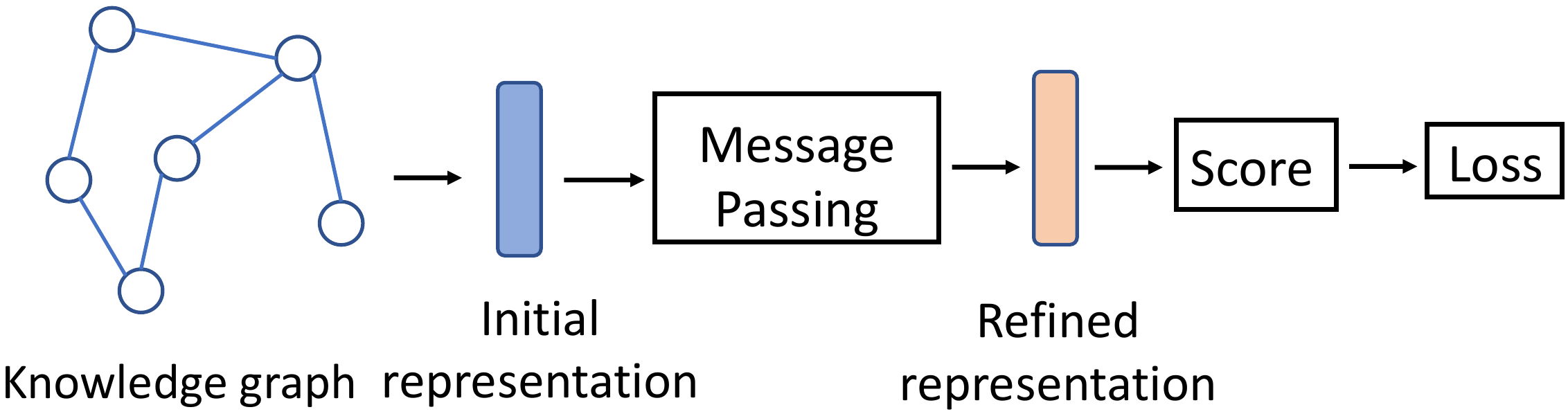}
  \caption{{A general MPNN framework for KGC. }}
  \label{fig:framework}
   \vspace{-0.2in}
\end{figure}

\vspace{-0.05in}
\subsection{MPNN-based KGC}\label{sec:gnn-kg}
Various MPNN-based models have been utilized for KGC by learning representations for the entities and relations of KGs. The learnt representations are then used as input to a scoring function $f(\cdot)$. Next, we first introduce MPNN models specifically designed for KG. Then, we introduce scoring functions. Finally, we describe the training process, including loss functions. 

\vspace{-0.05in}
\subsubsection{MPNNs for learning KG representations}
\label{sec:aggregation}
KGs can be naturally treated as graphs with triplets being the relational edges. When MPNN models are adapted to learn representations for KGs, the MP process in the MPNN layers is tailored for handling such relational data (triplets). In this paper, we investigate three representative MPNN-based models for KGC, i.e., \textbf{CompGCN} \cite{VashishthSNT20}, \textbf{RGCN} \cite{schlichtkrull2018modeling} and \textbf{KBGAT}~\citep{nathani2019learning}, which are most widely adopted.
As in standard MPNN models, these models stack multiple layers to iteratively aggregate information throughout the KG. Each intermediate layer takes the output from the previous layer as the input, and the final output from the last layer serves as the learned embeddings. In addition to entity embeddings, some MPNN-based models also learn relation embeddings. For a triplet $(h,r,t)$, we use ${\bf x}^{(k)}_h$, ${\bf x}^{(k)}_r$, and ${\bf x}^{(k)}_t$ to denote the head, relation, and tail embeddings obtained after the $k$-th layer. Specifically, the input embeddings of the first layer ${\bf x}^{(0)}_h$ , ${\bf x}^{(0)}_r$ and ${\bf x}^{(0)}_t$ are randomly initialized. RGCN aggregates neighborhood information with the relation-specific transformation matrices. CompGCN
defines direction-based transformation matrices and introduces relation embeddings to aggregate the neighborhood information. It introduces the composition operator to combine the embeddings to leverage the entity-relation information. KBGAT proposes attention-based aggregation process by considering both the entity embedding and relation embedding. More details about the MP process for CompGCN, RGCN and KBGAT can be found in {\bf Appendix~\ref{app:aggreation}}. For MPNN-based models with $K$ layers, we use ${\bf x}^{(K)}_h$, ${\bf x}^{(K)}_r$, and ${\bf x}^{(K)}_t$ as the final output embeddings and denote them as ${\bf x}_h$ , ${\bf x}_r$, and ${\bf x}_t$ for the simplicity of notations. Note that RGCN does not involve relation embedding ${\bf x}_r$ in the MP process, which will be randomly initialized if required by the scoring function.

\subsubsection{Scoring functions}\label{sec:scoring_function}

After obtaining the final embeddings from the MP layers, they are utilized as input to the scoring function $f$. Various scoring functions can be adopted. Two widely used scoring functions are DistMult~\citep{YangYHGD14a} and ConvE~\citep{ettmersMS018}. More 
specifically, RGCN adopts DistMult. In CompGCN, both scoring functions are investigated and ConvE is shown to be more suitable in most cases. Hence, in this paper, we use ConvE as the default scoring function for CompGCN. See {\bf Appendix~\ref{app:score}} for more scoring function details.





\subsubsection{Training MPNN-based models for KGC}\label{sec:training}


To train the MPNN model, the KGC task is often regarded as a binary classification task to differentiate the true triplets from the randomly generated ``fake'' triplets.
During training, all triplets in $\mathcal{D}_{train}$ and the corresponding inverse triplets $\mathcal{D}'_{train} = \{(t,r_{in},h)| (h,r,t) \in \mathcal{D}_{train}\}$ are treated as positive samples, where $r_{in}$ is the inverse relation of $r$. The final positive sample set can be denoted as $\mathcal{D}^{*}_{train} = \mathcal{D}_{train} \bigcup \mathcal{D}'_{train}$. Negative samples are generated by corrupting the triplets in $\mathcal{D}^{*}_{train}$. 
Specifically, for a triplet $(e_1,rel,e_2)\in \mathcal{D}^{*}_{train}$, we corrupt it by replacing its tail entities with other entities in the KG. 
More formally, the set of negative samples corresponding to the triplet $(e_1,rel,e_2)$ is denoted as:
$\mathcal{C}_{(e_1,rel,e_2)} = \{(e_1,rel,e_2')| e_2' \in \mathcal{V}, e_2' \neq e_2\}$ where $\mathcal{V}$ is the set of entities in KG. CompGCN uses $\mathcal{C}_{(e_1,rel,e_2)}$ as the negative samples. However, not all negative samples are utilized for training the RGCN model. Instead, for each positive sample triplet in $\mathcal{D}^{*}_{train}$, they adopt negative sampling to select $10$ such samples from  $\mathcal{C}_{(e_1,rel,e_2)}$, and use only these for training. Also, for RGCN, any relation ${r}$ and its inverse relation $r_{in}$ share the same diagonal matrix for DistMult in Eq.~\eqref{eq:dist} in {\bf Appendix~\ref{app:score}}. Both CompGCN and RGCN adopt the Binary Cross-Entropy (BCE) loss. More details are given in {\bf Appendix~\ref{app:loss}}.



\begin{figure*}[t]
\begin{center}
 \centerline{
{\subfigure[]
{\includegraphics[width=0.3\linewidth]{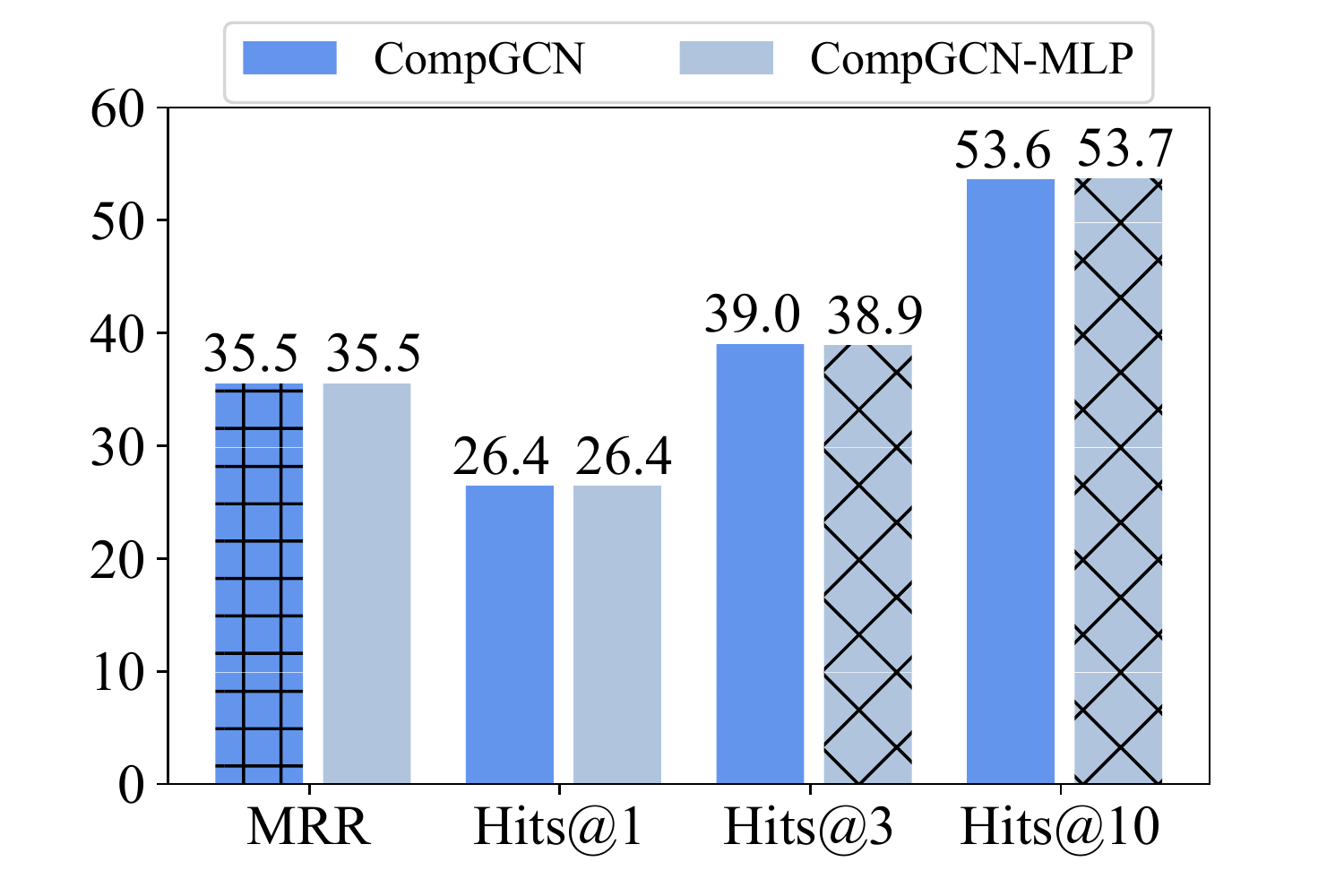} }}

{\subfigure[]
{\includegraphics[width=0.3\linewidth]{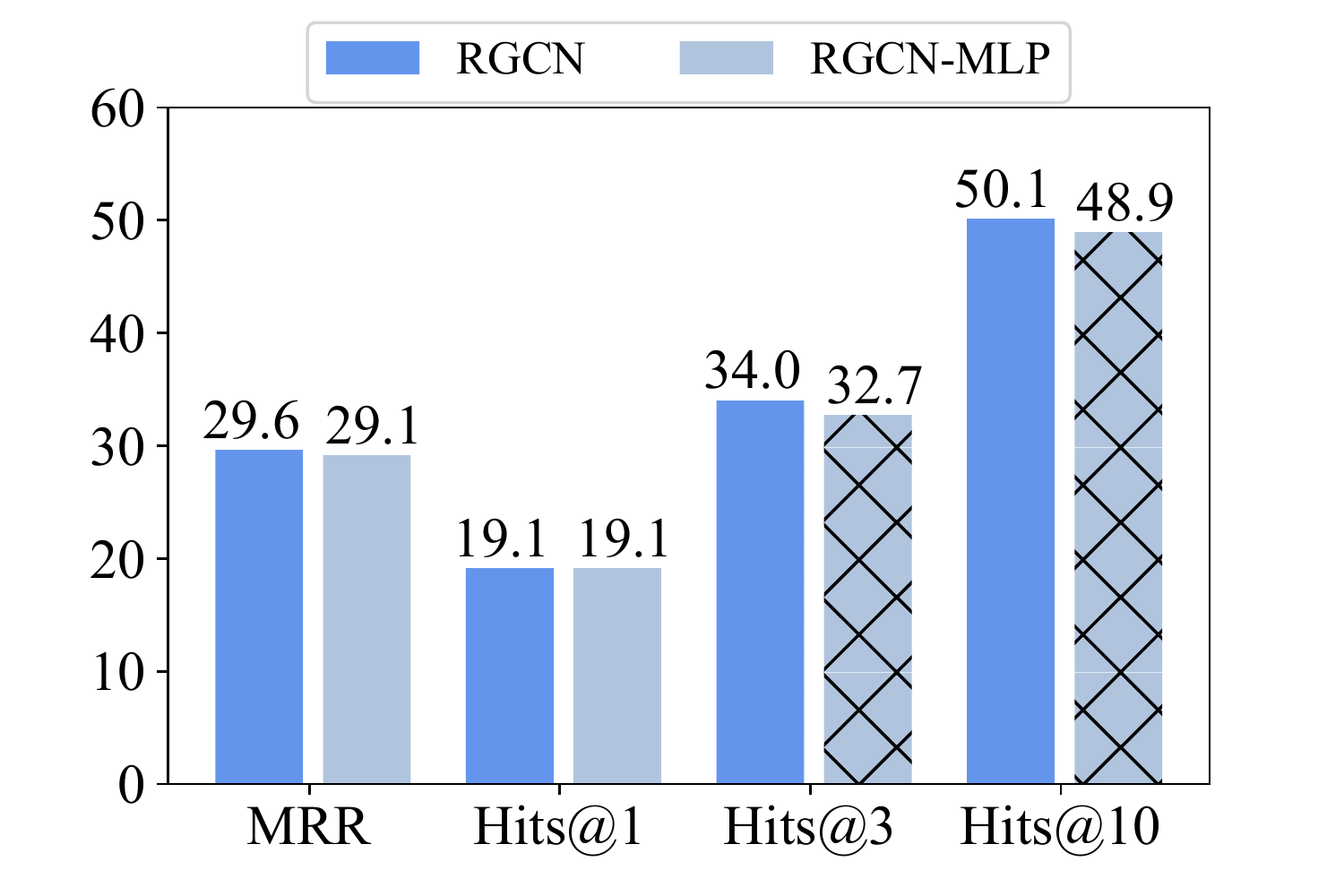} }}

{\subfigure[]
{\includegraphics[width=0.3\linewidth]{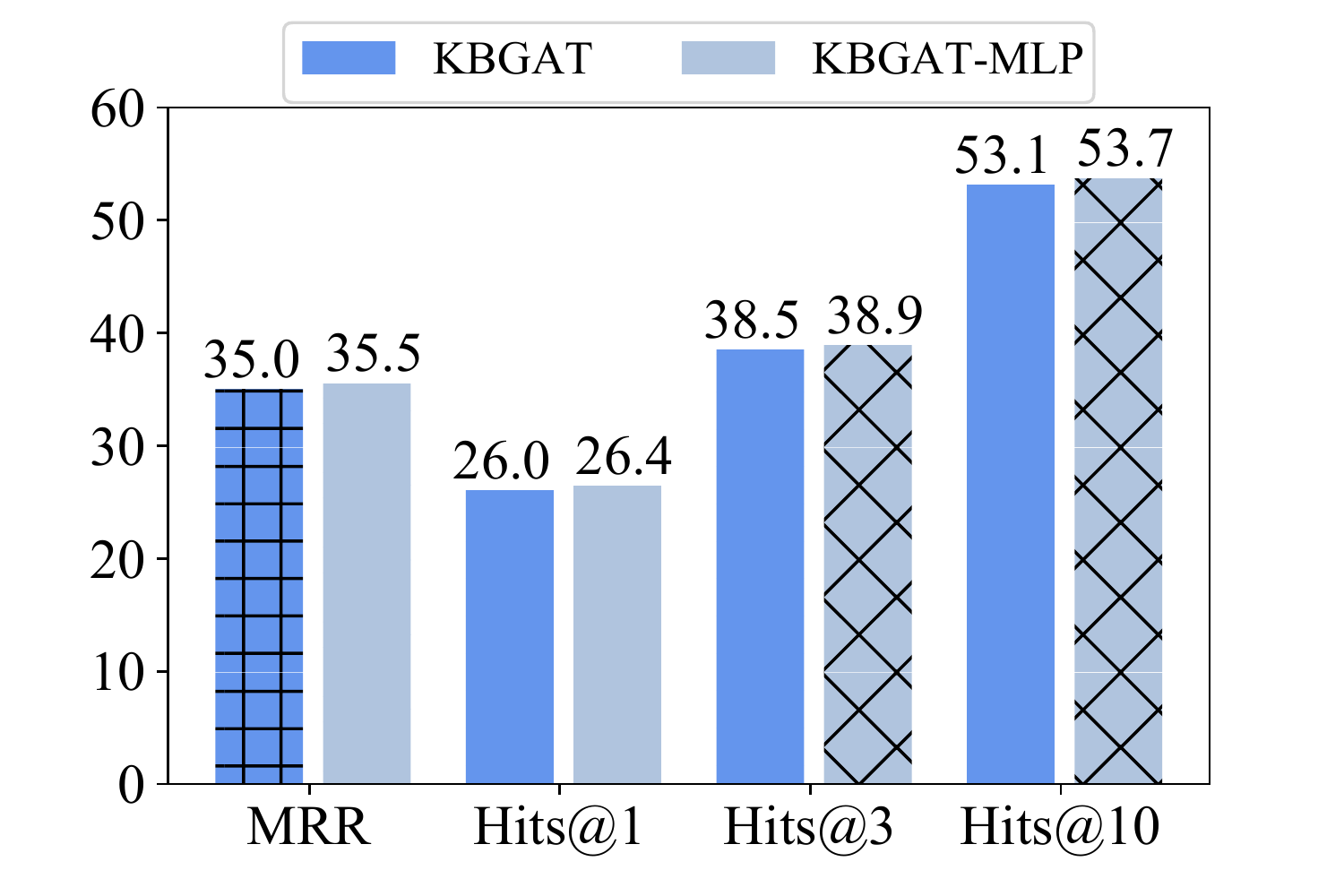} }}
}

\vspace{-0.15in}
\caption{KGC results (\%) of CompGCN/CompGCN-MLP, RGCN/RGCN-MLP, and KBGAT/KBGAT-MLP on \fbt. The MLP counterparts achieve compare performance as the corresponding MPNN models.}
\vspace{-0.4in}
\label{fig:com_finding}
\end{center}

\end{figure*}








\subsubsection{Major differences between MPNNs}\label{sec:diff}


We demonstrate an overview of MPNN-based model frameworks for the KGC task in Figure~\ref{fig:framework}. Specifically, the framework consists of several key components including the MP (introduced in Section~\ref{sec:aggregation}), the scoring function (\ref{sec:scoring_function}), and the loss function (~\ref{sec:training}). Training can typically be conducted end-to-end. Both RGCN and CompGCN follow this framework with various designs in each component. We provide a more detailed comparison about them later in this section. However, KBGAT  adopts a two-stage training process, which separates the training of the MP process (representation learning) and the scoring function. KBGAT achieves strong performance as reported in the original paper~\cite{nathani2019learning}, which was later attributed to a test leakage issue~\cite{sun2020re}. After addressing this test leakage issue, we found that fitting KBGAT into the general framework described in Figure~\ref{fig:framework} leads to much higher performance than training it with the two-stage process (around $10\%$ improvement on \fbt). Hence, in this paper, we conduct analyses for KBGAT by fitting its MP process (described in {\bf Appendix~\ref{app:aggreation}}) into the framework described in Figure~\ref{fig:framework}.

We summarize the major differences between RGCN, CompGCN, and KBGAT across three major components:
\textbf{(1)} \textbf{Message Passing.} Their MP processes are different as described in Section~\ref{sec:aggregation} and detailed in {\bf Appendix~\ref{app:aggreation}}.
\textbf{(2)} \textbf{Scoring Function.} They adopt different scoring functions. RGCN adopts the DistMult scoring function while CompGCN achieves best performance with ConvE. Thus, in this paper, we use ConvE as its default scoring function. For KBGAT, we adopt ConvE as its default scoring function.
\textbf{(3)} \textbf{Loss Function.} As described in Section~\ref{sec:training}, CompGCN utilizes all entities in the KG as negative samples for training, while RGCN adopts a negative sampling strategy. For KBGAT, we also utilize all entities to construct negative samples, similar to CompGCN.

%% file: 4_primary_findings.tex
\section{What Matters for  MPNN-based KGC?}

Recent efforts in adapting MPNN models for KG mostly focus on designing more sophisticated MP components to better handle multi-relational edges. These recently proposed MPNN-based methods have reported {\it strong performance} on the KGC task. Meanwhile,
RGCN, CompGCN and KBGAT achieve different performance. Their strong performance compared to traditional embedding based models and their performance difference are widely attributed to the MP  components~\cite{schlichtkrull2018modeling,VashishthSNT20,nathani2019learning}. However, as summarized in Section~\ref{sec:diff}, they differ from each other in several ways besides MP; little attention has been paid to understand how each component affects these models. Thus, what truly matters for MPNN-based KGC performance is still unclear. To answer this question, we design careful experiments to ablate the choices of these components in RGCN, CompGCN and  KBGAT to understand their roles, across multiple datasets. All reported  results are  mean and standard deviation over three seeds. 
Since MP is often regarded as the major contributor, we first investigate: is MP really helpful? Subsequently, we study the impact of the scoring function and the loss function.


\begin{table*}[!ht]
\centering
\vspace{-0.1in}
\caption{KGC results (\%) with various scoring functions. Models behave differently with different scoring functions. }
\vspace{-0.1in}
\begin{adjustbox}{width =1 \textwidth}
\begin{tabular}{cc|cccc|cccc |cccc }
\toprule
   & & \multicolumn{4}{c|}{\fbt} & \multicolumn{4}{c|}{\wnrr} &\multicolumn{4}{c}{\nell}    \\ 
         &   & MRR  & Hits@1   & Hits@3  & Hits@10  & MRR  & Hits@1  & Hits@3  & Hits@10   &MRR   & Hits@1   & Hits@3  & Hits@10 \\ \midrule

\multirow{2}{*}{CompGCN} &DistMult     & 33.7$\pm$ 0.1 & 24.7$\pm$0.1 &36.9$\pm$0.2 &51.5$\pm$0.2 &   42.9$\pm$0.1 & 39.0$\pm$0.1 & 43.9$\pm$0.1 & 51.7$\pm$0.3 &32.3$\pm$0.5&	24.3$\pm$0.6&	36.1$\pm$0.4&	47.4$\pm$0.2 \\ 
& ConvE& 35.5$\pm$0.1 &26.4$\pm$0.1 &39.0$\pm$0.2 & 53.6$\pm$0.3 &47.2$\pm$0.2 & 43.7$\pm$0.3 &48.5$\pm$0.3 &54.0$\pm$0.0 &38.1$\pm$0.4	&30.4$\pm$ 0.5&	42.2$\pm$0.3&	52.9$\pm$ 0.1\\  
\midrule
\multirow{2}{*}{RGCN}  &DistMult     & 29.6$\pm$0.3 &19.1$\pm$0.5 &34.0$\pm$ 0.2 &50.1$\pm$0.2   &43.0$\pm$0.2 & 38.6$\pm$0.3 &45.0$\pm$0.1 &50.8$\pm$0.3 &27.8$\pm$0.2 &	19.9$\pm$0.2&	31.4$\pm$0.0&	43.0$\pm$0.3 \\ 
& ConvE& 29.6$\pm$0.4 &20.3$\pm$ 0.4& 32.7$\pm$0.5& 47.9$\pm$0.6 &28.9$\pm$ 0.7& 17.4$\pm$ 0.8& 36.9$\pm$0.5& 48.8$\pm$0.5 &31.7$\pm$0.2 &	23.3$\pm$0.2 &	35.3$\pm$0.3& 48.5$\pm$ 0.2\\
\midrule
\multirow{2}{*}{KBGAT}  &DistMult   & 33.4$\pm$0.1&	24.5$\pm$0.1	& 36.6$\pm$0.1&	51.3$\pm$0.5 & 42.1$\pm$0.4&	38.7$\pm$0.4 &	43.1$\pm$ 0.6&	49.6$\pm$0.6& 33.0$\pm$0.2 &	25.5$\pm$ 0.1&	36.8$\pm$ 0.5& 47.3$\pm$0.5 \\
& ConvE& 35.0$\pm$0.3&	26.0$\pm$0.3 &	38.5$\pm$ 0.3&	53.1$\pm$ 0.3&46.4$\pm$0.2&	42.6$\pm$0.2&	47.9$\pm$0.3&53.9$\pm$0.2& 37.4$\pm$0.6&	29.7$\pm$0.7	&41.4$\pm$	0.8&52.0$\pm$0.4 \\
\bottomrule
\end{tabular}
\label{tab:inv_score}
\vspace{-0.1in}
\end{adjustbox}
\end{table*}

\begin{table*}[!ht]
\centering

\caption{KGC results  (\%) with various loss functions. The loss function significantly impacts model performance.
}
\begin{adjustbox}{width =1 \textwidth}
\begin{tabular}{cc|cccc|cccc  |cccc}
\toprule
   & & \multicolumn{4}{c|}{\fbt} & \multicolumn{4}{c|}{\wnrr} &\multicolumn{4}{c}{\nell}    \\ 
         &   & MRR  & Hits@1  & Hits@3 & Hits@10 & MRR & Hits@1 & Hits@3 & Hits@10 & MRR & Hits@1 & Hits@3 & Hits@10  \\ \midrule

\multirow{2}{*}{CompGCN} &with &31.5$\pm$0.1 &   {22.2}$\pm$0.1 & {34.8}$\pm$0.2 & {49.6}$\pm$0.2& {32.9}$\pm$ 0.9& {24.4}$\pm$ 1.5& {39.0}$\pm$0.5& {46.7}$\pm$0.4 & 32.0$\pm$0.2&	23.8$\pm$0.2&	35.7$\pm$0.1&	48.1$\pm$0.2\\
& w/o & 35.5$\pm$0.1 &26.4$\pm$0.1 &39.0$\pm$0.2 & 53.6$\pm$0.3 &47.2$\pm$0.2 & 43.7$\pm$0.3 &48.5$\pm$0.3 &54.0$\pm$0.0 &38.1$\pm$0.4	&30.4$\pm$ 0.5&	42.2$\pm$0.3&	52.9$\pm$ 0.1\\  
\midrule
\multirow{2}{*}{RGCN}  &with  &   29.6$\pm$0.3 &19.1$\pm$0.5 &34.0$\pm$0.2 &50.1$\pm$0.2   &43.0$\pm$0.2 & 38.6$\pm$0.3 &45.0$\pm$0.1 &50.8$\pm$0.3 &27.8$\pm$0.2 &	19.9$\pm$0.2&	31.4$\pm$0.0&	43.0$\pm$0.3 \\ 
& w/o &{33.4} $\pm$ 0.1& {24.3}$\pm$ 0.1& {36.7}$\pm$0.1 & {51.4}$\pm$ 0.2& 44.5$\pm$ 0.1& 40.9$\pm$0.1 & 45.5$\pm$ 0.1& 51.8$\pm$ 0.2 &34.6$\pm$ 0.6&	27.0$\pm$ 0.6&	38.3$\pm$ 0.6&	49.4$\pm$ 0.6\\
\midrule
\multirow{2}{*}{KBGAT}  &with  &30.1$\pm$ 0.3&	21.0$\pm$ 0.3&	33.2$\pm$ 0.4&	48.1$\pm$ 0.3& 30.1$\pm$ 0.2 &	18.6 $\pm$0.3&	37.8$\pm$0.3&	49.8$\pm$0.2& 32.6 $\pm$ 0.3&	24.3$\pm$ 0.3&	36.3$\pm$ 0.4& 	48.7$\pm$ 0.5\\
& w/o & 35.0$\pm$0.3&	26.0$\pm$0.3 &	38.5$\pm$ 0.3&	53.1$\pm$ 0.3&46.4$\pm$0.2&	42.6$\pm$0.2&	47.9$\pm$0.3&53.9$\pm$0.2& 37.4$\pm$0.6&	29.7$\pm$0.7	&41.4$\pm$	0.8&52.0$\pm$0.4 \\
\bottomrule
\end{tabular}
\end{adjustbox}
\vspace{-0.2in}
\label{tab:inv_loss}
\end{table*}

\subsection{Does Message Passing Really Help KGC?}
\label{sec:finding}




For RGCN and CompGCN, we follow the settings in the original papers to reproduce their reported performance. For KBGAT, we follow the same setting of CompGCN as mentioned in Section~\ref{sec:diff}. Specifically, we run these three models on datasets in their original papers. Namely,
we run RGCN on \fbt, \wn and \fb, CompGCN on \fbt and \wnrr, and  KBGAT on  \fbt, \wnrr and \nell. To understand the role of the MP component, we  keep other components untouched and replace their MP components with a simple MLP, which has the same number of layers and hidden dimensions with the corresponding MPNN-based models; note that since an MPNN layer is simply an aggregation over the graph combined with a feature transformation~\cite{ma2021unified}, replacing the MP component with MLP can also be achieved by replacing the adjacency matrix of the graph with an identity matrix. We denote the MLP models corresponding to RGCN, CompGCN and KBGAT as RGCN-MLP, CompGCN-MLP and KBGAT-MLP, respectively.
We present results for CompGCN,  RGCN and KBGAT\footnote[1]{We conduct a similar experiment using the setting in the original KBGAT paper.
 We find that KBGAT and KBGAT-MLP have similar performance on \fbt, \wnrr and \nell, which is consistent with Observation 1.} on the \fbt  in \figurename~\ref{fig:com_finding}. Due to the space limit, we present results on other datasets in {\bf Appendix~\ref{app:exp_gcn_vs_mlp}}.
We summarize the key observation from these figures: 

\begin{finding}\label{find:agg} 
The counterpart MLP-based models (RGCN-MLP, CompGCN-MLP and KBGAT-MLP) achieve comparable performance to their corresponding MPNN-based models on all datasets, suggesting that MP does not significantly improve model performance.
\label{finding}
\end{finding}
To further verify this observation, we investigate how the model performs when the graph structure utilized for MP is replaced by random generated graph structure. We found that the MPNN-based models still achieve comparable performance, which further verifies that the MP is not the major contributor. More details are in {\bf Appendix~\ref{app:exp_noise}.}

Moreover, comparing RGCN with CompGCN on \fbt in Figure~\ref{fig:com_finding}, we observe very different performances, while also noting that Observation~\ref{find:agg} clarifies that the difference in MP components is not the main contributor. This naturally raises a question: what are the important contributors? According to Section~\ref{sec:diff}, RGCN and CompGCN also adopt different scoring and loss functions, which provides some leads in answering this question. Correspondingly, we next empirically analyze the impact of the scoring and the loss functions with comprehensive experiments. Note that \fb and \wn suffer from the inverse relation leakage issue~\citep{toutanova2015observed,ettmersMS018}: a large number of test triplets can be obtained from inverting the triplets in the training set. Hence, to prevent these inverse relation leakage from affecting our studies, we conduct experiments on three datasets \nell, \fbt and \wnrr, where \fbt and \wnrr are the filtered versions of \fb and \wn after addressing these leakage issues.



\begin{table*}[!ht]
\centering
\caption{KGC results  (\%) with varying number of negative samples in the loss function. Generally,  utilizing $10$ negative samples is not enough. For different datasets and methods, the optimal number of negative samples varies.}
\begin{adjustbox}{width =1 \textwidth}
\begin{tabular}{cc|cccc|cccc  |cccc }
\toprule
   &  & \multicolumn{4}{c|}{\fbt} & \multicolumn{4}{c}{\wnrr} &\multicolumn{4}{c}{\nell}    \\ 
         &  \#Neg & MRR  & Hits@1  & Hits@3 & Hits@10 & MRR & Hits@1 & Hits@3 & Hits@10 & MRR & Hits@1 & Hits@3 & Hits@10  \\ \midrule

\multirow{5}{*}{CompGCN} &10 &{31.5}$\pm$0.1 &   {22.2}$\pm$ 0.1& {34.8}$\pm$ 0.2& {49.6}$\pm$0.2 & {32.9}$\pm$ 0.9& {24.4}$\pm$1.5 & {39.0}$\pm$ 0.5& {46.7}$\pm$ 0.4 &32.0$\pm$ 0.2&	23.8$\pm$ 0.2&	35.7$\pm$ 0.1&	48.1$\pm$0.2 \\
&50 &34.3$\pm$ 0.1&{24.7}$\pm$ 0.1& {38.1}$\pm$ 0.1 &{53.0}$\pm$ 0.1&  {40.0}$\pm$ 0.4& {33.0}$\pm$0.7 & {44.0}$\pm$0.2 & {51.6}$\pm$ 0.1& 37.2$\pm$ 0.9&	28.7$\pm$0.9	&41.6$\pm$ 0.9&	53.1$\pm$ 1.0\\
&200 &{35.3}$\pm$0.3  &{25.5}$\pm$ 0.1 &{39.2} $\pm$ 0.1&{53.8}$\pm$ 0.1 &{43.6}$\pm$ 0.5 & {39.2}$\pm$0.7  & {45.3}$\pm$ 0.2 & {52.3}$\pm$  0.5& 39.2$\pm$ 0.3 &	31.0$\pm$ 0.3&	43.6$\pm$ 0.2&	54.3 $\pm$ 0.2\\
&0.5$N$& {34.6} $\pm$0.1 & {25.3}$\pm$0.1 & {38.3}$\pm$ 0.1& {52.7}$\pm$0.1 & {44.0}$\pm$ 0.5& {40.6}$\pm$0.6 & {45.1}$\pm$ 0.6& {50.9}$\pm$ 0.3 &40.7$\pm$0.2 &	33.4$\pm$ 0.2&	44.4$\pm$ 0.3&	54.7$\pm$0.2 \\  
& $N$& {34.2}$\pm $0.1 &{25.0}$\pm$0.2  & {37.9}$\pm$ 0.3 & {52.2}$\pm$ 0.1 &{44.0}$\pm$0.3 & {40.7} $\pm$0.2 & {45.1}$\pm$ 0.3& {50.8}$\pm$0.5 & 40.3$\pm$ 0.5&	33.0$\pm$ 0.5&	44.0$\pm$ 0.5&	        54.4$\pm$0.4\\
\midrule
\multirow{5}{*}{RGCN}  &10    & 29.6$\pm$ 0.3 &19.1$\pm$0.5 &34.0$\pm$ 0.2 &50.1$\pm$0.2   &43.0$\pm$0.2 & 38.6$\pm$0.3 &45.0$\pm$0.1 &50.8$\pm$0.3 &27.8$\pm$0.2 &	19.9$\pm$0.2&	31.4$\pm$0.0&	43.0$\pm$0.3 \\
&50 &{32.5}$\pm$0.2 & {22.5}$\pm$0.3 & {36.7}$\pm$0.1 &{52.0}$\pm$ 0.4 &43.9$\pm$ 0.1& 39.6$\pm$ 0.1 &45.6$\pm$0.2 &51.8$\pm$ 0.2  & 29.6$\pm$0.3	&21.7$\pm$ 0.3&	33.2$\pm$ 0.3&	44.6$\pm$0.3\\
&200 & {33.2}$\pm$0.1 &{23.2}$\pm$ 0.1 &{37.6}$\pm$0.2 &{52.2}$\pm$0.3 &44.1$\pm$0.2 &39.9$\pm$ 0.4&45.7$\pm$0.2 &52.0$\pm$ 0.2& 30.0$\pm$0.3 &21.7$\pm$0.2 &	34.3$\pm$0.4	&45.7$\pm$0.3 \\    
&0.5$N$ & {33.3}$\pm$0.2 &{24.3}$\pm$0.3 &{36.9}$\pm$0.1 &{50.9}$\pm$0.2 &44.4$\pm$0.2 &40.7$\pm$ 0.3 &45.5$\pm$0.2 &52.0$\pm$ 0.3& 33.6$\pm$0.3 &	26.7$\pm$ 0.3&	37.2$\pm$0.2 &	46.7$\pm$0.2\\    
& $N$& {33.0}$\pm$0.4 & {23.9}$\pm$0.6 & {36.5}$\pm$ 0.3& {50.5}$\pm$0.3 & 44.5$\pm$0.2 & 40.6$\pm$0.2 & 45.8$\pm$ 0.2& 52.4$\pm$0.3 & 33.7$\pm$ 0.0&	26.9$\pm$ 0.0&	37.0$\pm$ 0.1&	46.4$\pm$ 0.2\\
\midrule
\multirow{5}{*}{KBGAT} & 10 & 30.1$\pm$0.3&	21.0$\pm$ 0.3&	33.2$\pm$ 0.4&	48.1$\pm$ 0.3& 30.1$\pm$ 0.2 &	18.6 $\pm$0.3&	37.8$\pm$0.3&	49.8$\pm$0.2& 32.6 $\pm$0.3&	24.3$\pm$0.3&	36.3$\pm$0.4& 	48.7$\pm$0.5\\

&50 &33.6$\pm$0.2	&24.2$\pm$0.3&	37.3$\pm$0.3	&51.9$\pm$0.2 &35.6$\pm$0.5 &	25.7$\pm$0.9&	42.6$\pm$0.3&	51.3$\pm$0.1 & 37.4$\pm$0.3&	29.0$\pm$0.3&	41.9$\pm$0.2 &	53.6$\pm$0.3\\

&200 & 34.7$\pm$0.2 &	25.1$\pm$0.2&	38.8$\pm$0.2&	53.3$\pm$0.2 &39.6$\pm$2.3&	32.7$\pm$3.7 &	43.4$\pm$0.8&	51.6$\pm$ 0.4 &39.2$\pm$0.2 &31.0$\pm$ 0.3&	43.6$\pm$0.1&	54.3$\pm$0.1 \\
&0.5$N$ &34.0$\pm$ 0.1&	24.7$\pm$ 0.1&	37.7$\pm$0.1&	52.2$\pm$0.1& 44.3$\pm$0.1 &	40.8$\pm$ 0.3&	45.5$\pm$ 0.2&	51.1$\pm$ 0.3& 40.1$\pm$0.1&	33.2$\pm$0.1&	43.5$\pm$0.2&	53.2$\pm$0.2 \\
			
& $N$ & 33.6$\pm$0.1 &	24.4$\pm$0.2 &	37.3$\pm$0.2	&52.0$\pm$0.3 & 43.8$\pm$ 0.9&	40.1$\pm$ 1.4&	45.3$\pm$ 0.5&	51.1$\pm$ 0.4& 39.6$\pm$ 0.2&	32.8$\pm$ 0.3&	43.0$\pm$0.3	&52.9$\pm$0.1  \\
\bottomrule
\end{tabular}
\end{adjustbox}
\vspace{-0.2in}
\label{tab:inv_negativesample}
\end{table*}

\subsection{Scoring Function Impact}
\label{sec:score}




 
Next, we investigate the impact of the scoring function on CompGCN, RGCN 
and KBGAT while fixing their loss function and experimental setting mentioned in Section~\ref{sec:diff}. 
The KGC results  are shown in Table~\ref{tab:inv_score}. 
In the original setting, CompGCN and KBGAT use ConvE as the scoring function while RGCN adopts DistMult. In Table~\ref{tab:inv_score}, we further present the results of CompGCN and KBGAT with DistMult and RGCN with ConvE. Note that we only make changes to the scoring function, while fixing all the other settings. Hence, in Table~\ref{tab:inv_score}, we still use RGCN, CompGCN and KBGAT to differentiate these three models but use DistMult and ConvE to indicate the specific scoring functions adopted. 

From this table, we have several observations:
\textbf{(1)} In most cases, CompGCN, RGCN and KBGAT behave differently when adopting different scoring functions. For instance, CompGCN and KBGAT achieve better performance when adopting ConvE as the scoring function in three datasets. RGCN with DistMult performs similar to that with ConvE on \fbt. However, it dramatically outperforms RGCN with ConvE on \wnrr and \nell. This indicates that the choice of scoring functions has strong impact on the performance, and the impact is dataset-dependent.
\textbf{(2)} Comparing CompGCN (or KBGAT) with RGCN on \fbt, even if the two methods adopt the same scoring function (either DistMult or ConvE), they still achieve quite different performance. On the \wnrr dataset, the observations are more involved. The two methods achieve similar performance when DistMult is adopted but behave quite differently with ConvE. Overall, these observations indicate that the scoring function is not the only factor impacting the model performance. 
    

\subsection{Loss Function Impact}
\label{sec:loss}
In this subsection, we investigate the impact of the loss function on these three methods while fixing the  scoring function and other experimental settings. As introduced in Section~\ref{sec:training}, in the original settings,  CompGCN, RGCN and KBGAT adopt the BCE loss. The major difference in the loss function is that CompGCN and  KBGAT utilize all negative samples while RGCN adopts a sampling strategy to randomly select $10$ negative samples for training. For convenience, we use \emph{w/o sampling} and \emph{with sampling} to denote these two settings and investigate how these two settings affect the model performance. 
\subsubsection{Impact of negative sampling}
To investigate the impact of negative sampling strategy, we also run CompGCN and KBGAT under the \emph{with sampling} setting (i.e., using $10$ negative samples as the original RGCN), and RGCN under the \emph{w/o sampling} setting. The results are shown in Table~\ref{tab:inv_loss}, where we use ``with'' and ``w/o'' to denote these two settings. From Table~\ref{tab:inv_loss}, we observe that RGCN, CompGCN and KBGAT achieve stronger performance under the ``\emph{w/o sampling}'' setting on three datasets. Specifically, the performance of CompGCN dramatically drops by $30.3\%$ from $47.2$ to $32.9$  when adopting the sampling strategy, indicating that the sampling strategy significantly impacts model performance. Notably, only using $10$ negative samples proves insufficient. Hence, we further investigate how the number of negative samples affects the model performance in the following subsection.

\subsubsection{Impact of number of negative samples}
In this subsection, we investigate how the number of negative samples affects performance under the ``\emph{with sampling}'' setting for both methods. We run RGCN,  CompGCN and KBGAT with varyinng numbers of negative samples. Following the settings of scoring functions as mentioned in Section ~\ref{sec:diff}., we adopt DistMult for RGCN and ConvE for CompGCN and KBGAT as scoring functions. Table~\ref{tab:inv_negativesample} shows the results and \#Neg is the number of negative samples. Note that in Table~\ref{tab:inv_negativesample}, $N$ denotes the number of entities in a KG, thus $N$ differs across the datasets. In general, increasing the number of negative samples from $10$ to a larger number is helpful for all methods. This partially explains why the original RGCN typically under-performs CompGCN and KBGAT. On the other hand, to achieve strong performance, it is not necessary to utilize all negative samples; for example, on \fbt, CompGCN achieves the best performance when the number of negative samples is $200$; this is advantageous, as using all negative samples is more expensive.  In short, carefully selecting the negative sampling rate for each model and dataset is important.

\vspace{-0.1in}
\section{KGC without Message Passing}
\vspace{-0.1in}
It is well known that MP is the key bottleneck for scaling MPNNs to large graphs \cite{jin2021graph, zhang2021graphless, zhao2021stars}. Observation~\ref{find:agg} suggests that the MP may be not helpful for KGC. Thus, in this section, we investigate if we can develop MLP-based methods (without MP) for KGC that can achieve comparable or even better performance than existing MPNN methods. Compared with the MPNN models, MLP-based methods enjoy the advantage of being more efficient during training and inference, as they do not involve expensive MP operations. We present the time complexity in {\bf Appendix ~\ref{app:time}}. 
The scoring and loss functions play an important role in MPNN-based methods. Likewise, we next study the impact of the scoring and loss functions on MLP-based methods. 

\begin{table*}[!ht]
\vspace{-0.1in}
\centering
\caption{KGC results  (\%) of MLP-based methods with different combinations of scoring and loss functions. Both the scoring and loss functions impact the performance of MLP-based models. }
\vspace{-0.1in}
\begin{adjustbox}{width = 1\textwidth}
\begin{tabular}{cc|cccc|cccc |cccc }
\toprule
   & & \multicolumn{4}{c|}{\fbt} & \multicolumn{4}{c}{\wnrr} & \multicolumn{4}{c}{\nell}    \\ 
         & \#Neg  & MRR  & Hits@1  & Hits@3 & Hits@10 & MRR & Hits@1 & Hits@3 & Hits@10  & MRR  & Hits@1  & Hits@3 & Hits@10    \\ \midrule

\multirow{5}{*}{DistMult} &10 &{29.1}$\pm$0.3 &   {19.1}$\pm$0.4 & {32.7}$\pm$0.4 & {48.9}$\pm$0.3 & \textbf{44.0$\pm$ 0.0}& {39.5$\pm$}0.5 & \textbf{45.7$\pm$ 0.3} &\textbf{51.9$\pm$0.5} & 27.5$\pm$0.2&	20.0$\pm$0.1&	30.9$\pm$0.1&	42.0$\pm$0.3\\

&50 &{31.3$\pm$}0.2&{21.2}$\pm$0.3 & {35.4}$\pm$0.3 &{51.1}$\pm$0.1 &  {42.5}$\pm$ 0.4& {38.3}$\pm$0.5 & {43.9}$\pm$0.4 & {51.1}$\pm$0.2  & 27.5$\pm$ 0.4&	19.4$\pm$0.6&	31.7$\pm$0.3&	42.7$\pm$0.1\\

&200 &{32.5}$\pm$0.3 &{22.3}$\pm$0.3 &\textbf{37.1$\pm$0.3} & \textbf{52.0$\pm$ 0.2}& {41.5}$\pm$ 0.5& {37.6}$\pm$0.5 & {42.6}$\pm$ 0.6& {49.6} $\pm$ 0.6& 29.1$\pm$0.2 &	21.1$\pm$0.2 &	33.2$\pm$0.0 &	43.9 $\pm$0.3\\

&0.5$N$& {32.8}$\pm$0.2 & {23.4}$\pm$0.3 & {36.8}$\pm$ 0.1& {50.7}$\pm$ 0.1& {41.3}$\pm$0.6& {38.0}$\pm$0.4 & {42.4}$\pm$0.7 & {48.5}$\pm$1.5 &{31.5}$\pm$ 0.2&	23.9$\pm$0.2	&35.5$\pm$ 0.2&	45.3$\pm$0.3\\

& $N$& {32.7}$\pm$ 0.1 &{23.5} $\pm$0.1 & {36.5}$\pm$0.1  & {50.4}$\pm$ 0.1 & {41.4}$\pm$0.2 & {38.4}$\pm$0.2 & {42.2}$\pm$0.2 & {47.7}$\pm$0.2 &31.0$\pm$0.1	&23.4$\pm$0.3&	34.8$\pm$0.3&	45.1$\pm$0.4 \\

&w/o & \textbf{33.4$\pm$}0.2 & \textbf{24.5 $\pm$ 0.2}& {36.6$\pm$ 0.2} & {51.1$\pm$ 0.2} &43.3$\pm$ 0.1& \textbf{39.9$\pm$ 0.1} & 44.6$\pm$0.2 &50.7 $\pm$0.9 & {\bf 32.8$\pm$0.2}&	{\bf 25.0$\pm$0.2}&	{ \bf 36.5 $\pm$ 0.3} &	{\bf 47.7 $\pm$ 0.3} \\
\midrule
\multirow{7}{*}{ConvE} &10 & {30.3} $\pm$0.4 & {21.1} $\pm$0.5 & {33.6} $\pm$0.4 & {48.6} $\pm$0.4 & 35.5 $\pm$5.8 & 27.6 $\pm$ 8.4& 40.5 $\pm$3.7 & 49.3 $\pm$ 1.2& 31.6 $\pm$0.6&	23.5 $\pm$0.5&	35.0 $\pm$ 0.7 &	47.2 $\pm$0.6\\

&50& {34.0}$\pm$ 0.3& 24.5$\pm$ 0.3& 37.9$\pm$ 0.2& 52.6$\pm$0.2 &42.1$\pm$0.1 & 36.3$\pm$0.2 &45.0$\pm$0.1 &52.4$\pm$0.1 & 37.0$\pm$ 0.3&	28.6$\pm$ 0.4&	41.4$\pm$0.3&	53.1$\pm$0.1 \\

&200 &{35.0$\pm$0.0} & {25.5}$\pm$0.1 & {39.0}$\pm$ 0.1& {53.4}$\pm$0.1 & 44.2$\pm$0.3 & 39.9$\pm$ 0.4& 45.7$\pm$ 0.3& 52.6$\pm$0.1 & 38.9$\pm$0.3 &	30.8$\pm$0.4 &	43.3$\pm$0.4 &	{\bf 54.0$\pm$0.1} \\

&500 & 35.3$\pm$0.0&25.7$\pm$0.0& \textbf{39.2$\pm$ 0.2} &53.6$\pm$0.2 & 44.5$\pm$0.3&40.6$\pm$0.4& 45.7$\pm$ 0.2& 52.3$\pm$0.2& 39.3$\pm$0.3 &	31.6$\pm$0.3	&{\bf 43.5 $\pm$0.4}&	53.6$\pm$0.3 \\

&0.5$N$ & {34.3}$\pm$0.1 &{25.0}$\pm$ 0.2& {38.1}$\pm$ 0.0& {52.4}$\pm$0.0 & {45.4}$\pm$0.2 & {41.8}$\pm$0.3& {46.4}$\pm$0.3 & {52.6}$\pm$0.2 & {\bf 40.0 $\pm$ 0.1}&{\bf	33.3$\pm$ 0.2}& 43.3$\pm$0.1 &	52.9$\pm$0.1\\

&$N$ & {34.0}$\pm$0.1  & {24.8}$\pm$0.2  & {37.7}$\pm$ 0.1 &{51.9}$\pm$0.1  & {45.4}$\pm$0.2  & {41.8}$\pm$0.2  &{46.5}$\pm$0.3  & {52.4} $\pm$ 0.1 & 39.7$\pm$0.2 &	33.0$\pm$ 0.1 &	43.1$\pm$0.2 &	52.5$\pm$ 0.1 \\

&w/o &\textbf{35.5 $\pm$ 0.2} & \textbf{26.4$\pm$0.2}& {38.9$\pm$0.2} & \textbf{53.7$\pm$0.1} & \textbf{47.3$\pm$ 0.1} & \textbf{43.7$\pm$0.2} & \textbf{48.8 $\pm$0.1}&  \textbf{54.4$\pm$0.1} & 38.1 $\pm$0.5&	30.4$\pm$0.5&	42.1$\pm$0.5&	52.5$\pm$0.5\\

\bottomrule
\end{tabular}
\end{adjustbox}
\label{tab:inv_mlp}
\end{table*}

\begin{table*}[!ht]
\centering
\vspace{-0.1in}
\caption{KGC results  (\%) of the ensembled MLP-based methods,  which outperform the MPNN-based models.}
\vspace{-0.1in}
\begin{adjustbox}{width =1 \textwidth}
\begin{tabular}{c|cccc|cccc |cccc  }
\toprule
    & \multicolumn{4}{c|}{\fbt} & \multicolumn{4}{c}{\wnrr} & \multicolumn{4}{c}{\nell}    \\ 
            & MRR  & Hits@1  & Hits@3 & Hits@10 & MRR & Hits@1 & Hits@3 & Hits@10  & MRR  & Hits@1  & Hits@3 & Hits@10 \\ \midrule
CompGCN     &    35.5$\pm$0.1 &26.4$\pm$0.1 &39.0$\pm$0.2 & 53.6$\pm$0.3 &47.2$\pm$0.2 & 43.7$\pm$0.3 &48.5$\pm$0.3 &54.0$\pm$0.0 &38.1$\pm$0.4	&30.4$\pm$ 0.5&	42.2$\pm$0.3&	52.9$\pm$ 0.1 \\
RGCN     &   29.6$\pm$ 0.3 &19.1$\pm$0.5 &34.0$\pm$ 0.2 &50.1$\pm$0.2   &43.0$\pm$0.2 & 38.6$\pm$0.3 &45.0$\pm$0.1 &50.8$\pm$0.3 &27.8$\pm$0.2 &	19.9$\pm$0.2&	31.4$\pm$0.0&	43.0$\pm$0.3 \\
 KBGAT& 35.0$\pm$0.3&	26.0$\pm$0.3 &	38.5$\pm$ 0.3&	53.1$\pm$ 0.3&46.4$\pm$0.2&	42.6$\pm$0.2&	47.9$\pm$0.3&53.9$\pm$0.2& 37.4$\pm$0.6&	29.7$\pm$0.7	&41.4$\pm$	0.8&52.0$\pm$0.4 \\
 \midrule
MLP-best &  {35.5 $\pm$ 0.2} & {26.4$\pm$0.2}& {38.9$\pm$0.2} & {53.7$\pm$0.1} & {47.3$\pm$ 0.1} & {43.7$\pm$0.2} & {48.8 $\pm$0.1}&  {54.4$\pm$0.1}& {40.0 $\pm$ 0.1}&{33.3$\pm$ 0.2}& 43.3$\pm$0.1 &	52.9$\pm$0.1\\
MLP-ensemble     &   \textbf{36.9 $\pm$0.2}& \textbf{27.5$\pm$0.2} & \textbf{40.8$\pm$0.2} & \textbf{55.4$\pm$0.1} & \textbf{47.7$\pm$0.3} & \textbf{43.9 $\pm$0.4}& \textbf{48.9$\pm$0.1} & \textbf{55.4$\pm$0.1}  &\textbf{41.7$\pm$0.2}&	\textbf{34.7$\pm$0.2}	&\textbf{45.1$\pm$0.0}&	\textbf{55.2$\pm$0.1} \\     

\bottomrule
\end{tabular}
\end{adjustbox}
\vspace{-0.2in}
\label{tab:mlp_ensemble}
\end{table*}

\subsection{MLPs with various scoring and loss}
\label{sec:mlp}
We investigate the performance of MLP-based models with different combinations of scoring and loss functions. Specifically, we adopt DistMult and ConvE as scoring functions. For each scoring function, we try both the \emph{with sampling} and \emph{w/o sampling} settings for the loss function. Furthermore, for the \emph{with sampling} setting, we vary the number of negative samples. The results of MLP-based models with different combinations are shown in Table~\ref{tab:inv_mlp}, which begets the following observations: 
\textbf{(1)} The results from Table~\ref{tab:inv_mlp} further confirm that the  MP component is unnecessary for KGC.
The MLP-based models can achieve comparable or even stronger performance than GNN models.
\textbf{(2)} Similarly, the scoring and the loss functions play a crucial role in the KGC performance, though dataset-dependent. For example, it is not always necessary to adopt the \emph{w/o} setting for strong performance: On the \fbt dataset, when adopting ConvE for scoring, the MLP-based model achieves comparable performance with $500$ negative samples; on \wnrr, when adopting DistMult for scoring, the model achieves best performance with $10$ negative samples; on \nell, when adopting ConvE for scoring, it achieves the best performance with $0.5N$ negative samples.

Given these observations, next we study a simple ensembling strategy to combine different MLP-based models, to see if we can obtain a strong and limited-complexity model which can perform well for various datasets, without MP. Note that ensembling MLPs necessitates training multiple MLPs, which introduces additional complexity. However, given the efficiency of MLP, the computational cost of ensembling is still acceptable. 





\subsection{Ensembling MLPs}\label{sec:ensemble}
According to Section~\ref{sec:mlp}, the performance of MLP-based methods is affected by the scoring function and the loss function, especially the negative sampling strategy. These models with various combinations of scoring function and loss functions can potentially capture important information from different perspectives. Therefore, an ensemble of these models could provide an opportunity to combine the information from various models to achieve better performance. Hence, we select some MLP-based models that exhibit relatively good performance on the validation set and ensemble them for the final prediction. Next, we briefly describe the ensemble process. These selected models are individually trained, and then assembled together for the inference process. Specifically, during the inference stage, to calculate the final score for a specific triplet $(h,r,t)$, we utilize each selected model to predict a score for this triplet individually and then add these scores to obtain the final score for this triplet. The final scores are then utilized for prediction. In this work, our focus is to show the potential of ensembling instead of designing the best ensembling strategies; hence, we opt for simplicity, though more sophisticated strategies could be adopted. We leave this to future work.  

We put the details of the  MLP-based models we utilized for constructing the ensemble model for these three datasets in {\bf Appendix~\ref{app:ensemble}}.
The results of these ensemble methods are shown in Table~\ref{tab:mlp_ensemble}, where we use MLP-ensemble to generally denote the ensemble model. Note that MLP-best in the table denotes the best performance from individual MLP-based methods from Table~\ref{tab:inv_mlp}. From the table, we can clearly observe that MLP-best can achieve comparable or even slightly better performance than  MPNN-based methods. Furthermore, the MLP-ensemble can obtain better performance than both the best individual MLP-based methods and the MPNN-based models, especially on \fbt and \nell. These observations further support that the MP component is not necessary. They also indicate that these scoring and loss functions are potentially complementary to each other, and as a result, even the simple ensemble method can produce better performance.

%% file: 5_discussion.tex
\section{Discussion}
\noindent\textbf{Key Findings:} \textbf{(1)} The MP component in MPNN-based methods does not significantly contribute to KGC performance, and MLP-based methods without MP can achieve comparable performance; \textbf{(2)} Scoring and the loss function design (i.e. negative sampling choices) play a much more crucial role for both MPNN-based and MLP-based methods; \textbf{(3)} The impact of these is significantly dataset-dependent; and \textbf{(4)} Scoring and the loss function choices are complementary, and simple strategies to combine them in MLP-based methods can produce better KGC performance.


\noindent\textbf{Practical Implications:} \textbf{(1)} MLP-based models do not involve the complex MP process and thus they are more efficient than the MPNN-based models \cite{zhang2021graphless}. Hence, such models are more scalable and can be applied to large-scale KGC applications for practical impact; \textbf{(2)} The simplicity and scalability of MLP-based models make ensembling easy, achievable and effective (Section~\ref{sec:ensemble}); and \textbf{(3)} The adoption of MLP-based models enables us to more conveniently apply existing techniques to advance KGC. For instance, Neural Architecture Search (NAS) algorithms~\cite{zoph2016neural} can be adopted to automatically search better model architectures, since NAS research for MLPs is much more extensive than for MPNNs.


\noindent\textbf{Implications for Future Research:} \textbf{(1)} Investigating better designs of scoring and loss functions are (currently) stronger levers to improve KGC. Further dedicated efforts are required for developing suitable MP operations in MPNN-based models for this task; \textbf{(2)} MLP-based models should be adopted as default baselines for future KGC studies.  This aligns with several others which suggest the underratedness of MLPs for vision-based problems \cite{liu2021pay, tolstikhin2021mlp}; \textbf{(3)} Scoring and loss function choices have complementary impact, and designing better strategies to combine them is promising; and \textbf{(4)} Since KGC is a type of link prediction, and many works adopt MPNN designs in important settings like ranking and  recommendations~\cite{ying2018graph,fan2019graph,wang2019kgat}, our work motivates a  pressing need to understand the role of MP components in these applications.


%% file: 2_related_work.tex
\section{Related Work}

There are mainly two types of 
GNN-based KGC models: MPNN-based models and path-based models. When adopting MPNNs for KG, recent efforts have been made to deal with the multi-relational edges in KGs by designing MP operations.  RGCN~\citep{schlichtkrull2018modeling} introduces the relation-specific transformation matrices. CompGCN~\citep{VashishthSNT20} integrates neighboring information based on entity-relation composition operations. KBGAT~\citep{nathani2019learning} learns attention coefficients to distinguish the role of entity in various relations. Path-based models learn pair-wise representations by aggregating the path information between the two nodes. NBFNet~\citep{zhu2021neural} integrates the information from all paths between the two nodes. RED-GNN~\citep{zhang2022knowledge} makes use of the dynamic programming and A$^{*}$Net~\citep{zhu2022learning} prunes paths by prioritizing important nodes and edges. In this paper, we focus on investigating how the MP component in the MPNNs affects their performance in the KGC task. Hence, we do not include these path-based models into the comparison. A concurrent work~\citep{zhang2022rethinking} has similar observations as ours. However, they majorly focus on exploring  how the MP component affects the performance. Our work provides a more thorough analysis on the major contributors for MPNN-based KGC models and proposes a strong ensemble model based upon the analysis.  




%% file: 6_conclusion.tex
\section{Conclusion}
In this paper, we surprisingly find that the MLP-based models are able to achieve competitive performance compared with three MPNN-based models (i.e., CompGCN, RGCN and KBGAT) across a variety of datasets. It suggests that the message passing operation in these  models is not the key component to achieve strong performance. To explore which components potentially contribute to the model performance, we conduct extensive experiments on other key components such as scoring function and loss function. We found both of them play crucial roles, and their impact varies significantly across datasets. Based on these findings, we further propose ensemble methods built upon MLP-based models, which are able to achieve even better performance than MPNN-based models. 

\section{Acknowledgements}
This research is supported by the National Science Foundation (NSF) under grant numbers CNS1815636, IIS1845081, IIS1928278, IIS1955285, IIS2212032, IIS2212144, IOS2107215, IOS2035472, and IIS2153326, the Army Research Office (ARO) under grant number W911NF-21-1-0198, the Home Depot, Cisco Systems Inc, Amazon Faculty Award, Johnson\&Johnson and SNAP.

%% file: appendix.tex
\newpage


\appendix

\section{Dataset}
\label{sec:data}
We use five well-known KG datasets -- Table \ref{table:datasets} details their statistics:
\begin{compactitem}[\textbullet]
    \item \textbf{FB15k}~\cite{BordesUGWY13} is a subset of the Freebase database~\cite{bollacker2008freebase} containing general facts. It is constructed by selecting a subset of entities that are both in the Wikilinks database\footnote{https://code.google.com/archive/p/wiki-links/} and Freebase. 
    \item \textbf{FB15k-237} \cite{toutanova2015representing,toutanova2015observed} is a subset of the FB15k which removes the inverse relations from \fb to prevent direct inference.
    \item \textbf{WN18} \cite{schlichtkrull2018modeling} is subset of the WordNet database \cite{fellbaum2010wordnet} which contains lexical relations between words.
    \item \textbf{WN18RR} \cite{ettmersMS018} is a subset of the WN18. WN18 contains triplets in the test set that are generated by inverting triplets from the training set. WN18RR is constructed by removing these triplets to avoid inverse relation test leakage.
    \item \textbf{NELL-995}~\citep{XiongHW17} is constructed from the 995-th iteration of the NELL system~\citep{carlson2010toward} which constantly extracts facts from  the web.
\end{compactitem}

\begin{table}[]
\centering
\footnotesize
 \caption{Data statistics for four datasets.}
 \begin{adjustbox}{width =0.5 \textwidth}
\begin{tabular}{lccccc}
\toprule
   Datasets& Entities & Relations & Train edges &Val. edges  &Test edges  \\ 
   \midrule
    FB15k-237  & 14,541& 237  &272,115 & 17,535&20,466 \\
     WN18RR &40,943 &11&86,835 & 3,034 &3,134 \\
     WN18& 40,943&18&141,442 & 5,000 & 5,000 \\
     FB15k & 14,951 & 1,345 &483,142 & 50,000 & 59, 071 \\
     NELL-995 &75.492 &200  & 126,176 & 13,912 & 14,125\\
  \bottomrule
\end{tabular}
 \label{table:datasets}
 \end{adjustbox}
\end{table}

\section{Evaluation Metrics}
\label{sec:metric}
We use the rank-based measures to evaluate the quality of the prediction including  Mean Reciprocal Rank (\textbf{MRR}) and \textbf{Hits@N}. Their detailed definitions are introduced below: 
\begin{compactitem}[\textbullet]
    \item Mean Reciprocal Rank (\textbf{MRR}) is the mean of the reciprocal predicted rank for the ground-truth entity over all triplets in the test set. A higher MRR indicates better performance. 
    \item \textbf{Hits@N} calculates the proportion of the groundtruth tail entities with a rank smaller or equal to $N$ over all triplets in the test set. Similar to MRR, a higher Hits@N indicates better performance.
\end{compactitem}
 These metrics are indicative, but they can be flawed when a tuple (i.e., $(h,r)$ or $(r,t)$) has multiple ground-truth entities which appear in either the training, validation or test sets. Following the filtered setting in previous works \cite{BordesUGWY13, schlichtkrull2018modeling, VashishthSNT20}, we remove the misleading entities  when ranking and report the filtered results. 

\section{Message Passing in MPNN-based KGC}
\label{app:aggreation}

For a general triplet $(h,r,t)$, we use ${\bf x}^{(k)}_h$ , ${\bf x}^{(k)}_r$, and ${\bf x}^{(k)}_t$ to denote the head, relation, and tail embeddings obtained after the $k$-th layer. Specifically, the input embeddings of the first layer ${\bf x}^{(0)}_h$ , ${\bf x}^{(0)}_r$ and ${\bf x}^{(0)}_t$ are randomly initialized. Next, we describe the information aggregation process in the $(k+1)$-th layer for the studied three MPNN-based models, i.e., CompGCN, RGCN and KBGAT. 

\begin{compactitem}[\textbullet]
\item \textbf{RGCN}~\cite{schlichtkrull2018modeling} aggregates neighborhood information with the relation-specific transformation matrices:
\begin{equation}
\label{eq:rgcn}
    \mathbf{x}_h^{(k+1)} = g ( \sum_{( r,t) \in \mathcal{N}_h}\frac{1}{c_{h,r}} \mathbf{W}_r^{(k)} \mathbf{x}_t^{(k)} + \mathbf{W}_o^{(k)} \mathbf{x}_h^{(k)}  )
\end{equation}
where $\mathbf{W}_o^{(k)} \in \mathbb{R}^{d_{k+1} \times d_k}$ and $\mathbf{W}_r^{(k)} \in \mathbb{R}^{d_{k+1} \times d_k}$ are learnable matrices. $\mathbf{W}_r^{(k)}$ corresponds to the relation $r$, $\mathcal{N}_h$ is the set of neighboring tuples $(r,t)$ for entity $h$, $g$ is a non-linear function, and $c_{h,r}$ is a normalization constant that can be either learned or predefined.

\item \textbf{CompGCN}~\cite{VashishthSNT20}
defines direction-based transformation matrices and introduce relation embeddings to aggregate the neighborhood information:
\begin{equation}
\label{eq:compgcn}
    \mathbf{x}_t^{(k+1)} = g \left( \sum_{(h, r) \in \mathcal{N}_t} \mathbf{W}_{\lambda(r)}^{(k)} \: \phi(\mathbf{x}_h^{(k)}, \mathbf{x}_r^{(k)}) \right),
\end{equation}
where $\mathcal{N}_t$ is the set of neighboring entity-relation tuples $(h,r)$ for entity $t$, $\lambda(r)$ denotes the direction of relations: original relation, inverse relation, and self-loop. $\mathbf{W}_{\lambda(r)}^{(k)} \in \mathbb{R}^{d_{k+1} \times d_k}$ is the direction specific learnable weight matrix in the $k$-th layer, and $\phi(\cdot)$ is the composition operator to combine the embeddings to leverage the entity-relation information. The composition operator $\phi(\cdot)$ is defined as the subtraction, multiplication, or cross correlation of the two embeddings \cite{VashishthSNT20}. CompGCN generally achieves best performance when adopting the cross correlation. Hence, in this work, we use the cross correlation as its default composition operation for our investigation. CompGCN updates the relation embedding through linear transformation in each layer, i.e., $\mathbf{x}_r^{(k+1)} = \mathbf{W}_{rel}^{(k)} \mathbf{x}_r^{(k)}$ where $\mathbf{W}_{rel}^{(k)}$ is the learnable weight matrix.

\item \textbf{KBGAT}~\citep{nathani2019learning}  proposes attention-based aggregation process by considering both the entity embedding and relation embedding:
\begin{equation}
\label{eq:kbgat}
    \mathbf{x}_h^{(k+1)} = g  \left( \sum_{(r,t)\in  \mathcal{N}_h} \alpha_{h,r,t}^{(k)} c_{h,r,t}^{(k)}\right)
\end{equation}
where $ c_{h,r,t}^{(k)} = {\bf W}_1^{(k)}[{\bf x}_h^{(k)} || {\bf x}_t^{(k)} || {\bf x}_r]$,  $||$ is the concatenation operation. Note that the relation embedding is randomly initilized and shared by all layers, i.e, ${x_r^{(k)}=x_r}$. The coefficient $\alpha_{h,r,t}^{(k)}$ is the attention score for $(h,r,t)$ in the $k$-th layer, which is formulated as follows:
\begin{equation}
\label{eq:kbgat_attention}
   \alpha_{h,r,t}^{(k)} = \frac{\exp(\text{LR}({\bf W}_2^{(k)} c_{h,r,t}^{(k)}))}{\sum_{(r,t')\in \mathcal{N}_h}\exp(\text{LR}({\bf W}_2^{(k)} c_{h,r,t'}^{(k)}))}
\end{equation}
where LR is the LeakyReLU function, 
 ${\bf W}_1^{(k)} \in \mathbb{R}^{d_{k+1}\times 3d_{k}}$,  ${\bf W}_2^{(k)} \in \mathbb{R}^{1 \times d_{k+1}}$ are two sets of learnable parameters.

\end{compactitem}
For GNN-based models with $K$ layers, we use ${\bf x}^{(K)}_h$, ${\bf x}^{(K)}_r$, and ${\bf x}^{(K)}_t$ as the final embeddings and denote them as ${\bf x}_h$ , ${\bf x}_r$, and ${\bf x}_t$ for the simplicity of notations. Note that RGCN does not involve $x_r$ in the aggregation component, which will be randomly initialized if required by the scoring function.

 \begin{table*}[!ht]

\caption{KGC results (\%) with random graph structure for message passing process. The MPNN-based models still can achieve comparable performance.}
\begin{adjustbox}{width =1 \textwidth}
\begin{tabular}{cc|llll|llll |llll }
\toprule
   & & \multicolumn{4}{c|}{\fbt} & \multicolumn{4}{c|}{\wnrr} &\multicolumn{4}{c}{\nell}    \\ 
         &   & MRR  & Hits@1  & Hits@3 & Hits@10 & MRR & Hits@1 & Hits@3 & Hits@10  &MRR  & Hits@1  & Hits@3 & Hits@10\\ \midrule

\multirow{2}{*}{CompGCN} &Original &35.5$\pm$0.1 &26.4$\pm$0.1 &39.0$\pm$0.2 & 53.6$\pm$0.3 &47.2$\pm$0.2 & 43.7$\pm$0.3 &48.5$\pm$0.3 &54.0$\pm$0.0 &38.1$\pm$0.4	&30.4$\pm$ 0.5&	42.2$\pm$0.3&	52.9$\pm$ 0.1 \\
& Random  & 35.3$\pm$0.1	&26.3$\pm$0.1&	38.7$\pm$0.1&		53.4 $\pm$0.2&47.3$\pm$0.0&			44.0$\pm$0.2&				48.5$\pm$0.2&				53.8$\pm$0.3 & 38.8$\pm$0.1	&			31.1$\pm$0.0&				42.8$\pm$0.1		&		53.3$\pm$0.1\\

\midrule
\multirow{2}{*}{RGCN} & Original&  29.6$\pm$0.3 &19.1$\pm$0.5 &34.0$\pm$ 0.2 &50.1$\pm$0.2   &43.0$\pm$0.2 & 38.6$\pm$0.3 &45.0$\pm$0.1 &50.8$\pm$0.3 &27.8$\pm$0.2 &	19.9$\pm$0.2&	31.4$\pm$0.0&	43.0$\pm$0.3 \\
& Random & 28.6$\pm$0.5&			18.8$\pm$0.5&				32.4$\pm$0.8	&			48.2$\pm$0.7 & 43.0$\pm$0.3	&			38.7$\pm$0.1		&		45.0$\pm$0.5&				50.8$\pm$0.6 & 27.7$\pm$0.2		&		19.6$\pm$0.2	&			31.4$\pm$0.3		&		43.3$\pm$0.2\\

\midrule
\multirow{2}{*}{KBGAT}  &  Original&35.0$\pm$0.3&	26.0$\pm$0.3 &	38.5$\pm$ 0.3&	53.1$\pm$ 0.3&46.4$\pm$0.2&	42.6$\pm$0.2&	47.9$\pm$0.3&53.9$\pm$0.2& 37.4$\pm$0.6&	29.7$\pm$0.7	&41.4$\pm$	0.8&52.0$\pm$0.4 \\
& Random & 35.6$\pm$0.1&	26.5$\pm$0.1&	39.0$\pm$0.2&	53.7$\pm$0.1 & 46.8$\pm$0.2&	43.2$\pm$0.5&	48.1$\pm$0.1&	53.8$\pm$0.1 & 38.2$\pm$0.3&	30.6$\pm$0.3&	42.1$\pm$0.4&	52.8$\pm$0.2\\

\bottomrule
\end{tabular}
\label{tab:noise_in_aggregation}

\end{adjustbox}
\end{table*}

\section{Scoring Function}
\label{app:score}
Two widely used scoring function are DistMult~\citep{YangYHGD14a} and ConvE~\citep{ettmersMS018}. The definitions of these scoring functions are as follows.
\begin{align}
   & f^{{DistMult}}(h,r,t) = \mathbf{x}_h \mathbf{R}_r\mathbf{x}_t \label{eq:dist}\\
    &    f^{ConvE}(h,r,t) = g(\text{vec}(g([\overline{\mathbf{x}_h}||\overline{\mathbf{x}_r}]\ast \omega))\mathbf{W})
    \mathbf{x}_t \label{eq:conve}
\end{align}
\noindent $\mathbf{R}_r \in \mathbb{R}^{d_k \times d_k}$ in Eq.~\eqref{eq:dist} is a diagonal matrix corresponding to the relation $r$. In Eq.~\eqref{eq:conve}, $\overline{\mathbf{x}_h}$ denotes a 2D-reshaping of $\mathbf{x}_h$, $\omega$ is the convolutional filter, and $\mathbf{W}$ is the learnable matrix. vec($\cdot$) is an operator to reshape a tensor into a vector. $||$ is the concatenation operator. ConvE feeds the stacked 2D-reshaped head entity embedding and relation embedding into convolution layers. It is then reshaped back into a vector that multiplies the tail embedding to generate a score.

For DistMult, there are different ways to define the diagonal matrix ${\bf R}_r$: For example, in RGCN, the diagonal matrix is randomly initialized for each relation $r$, while CompGCN defines the diagonal matrix by diagonalizing the relation embedding ${\bf x}_r$.

\section{Loss Function}
\label{app:loss}
We adopt the Binary cross-entropy (BCE) as the loss function, which can be modeled as follows:
\begin{align}
  \nonumber \mathcal{L} = - \sum_{(e_1,rel,e_2) \in \mathcal{D}^{*}_{train}}\bigg( \log \sigma(f(e_1,rel,e_2))+ \\
   \sum_{ (e_1,rel,e_2')\in \mathcal{C}_{(e_1,rel,e_2)}} \log (1-\sigma(f(e_1,rel,e_2')))  \bigg).
    \label{Eq:loss}
\end{align}
where $f(\cdot)$ is the scoring function defined in the appendix~\ref{app:score}, and $\sigma$ is the sigmoid function.

\section{Does Message Passing Really Help KGC? }
\label{app:exp_gcn_vs_mlp}

\begin{figure}[t]
\begin{center}
 \centerline{

{\subfigure[\wnrr]
{\includegraphics[width=0.8\linewidth]{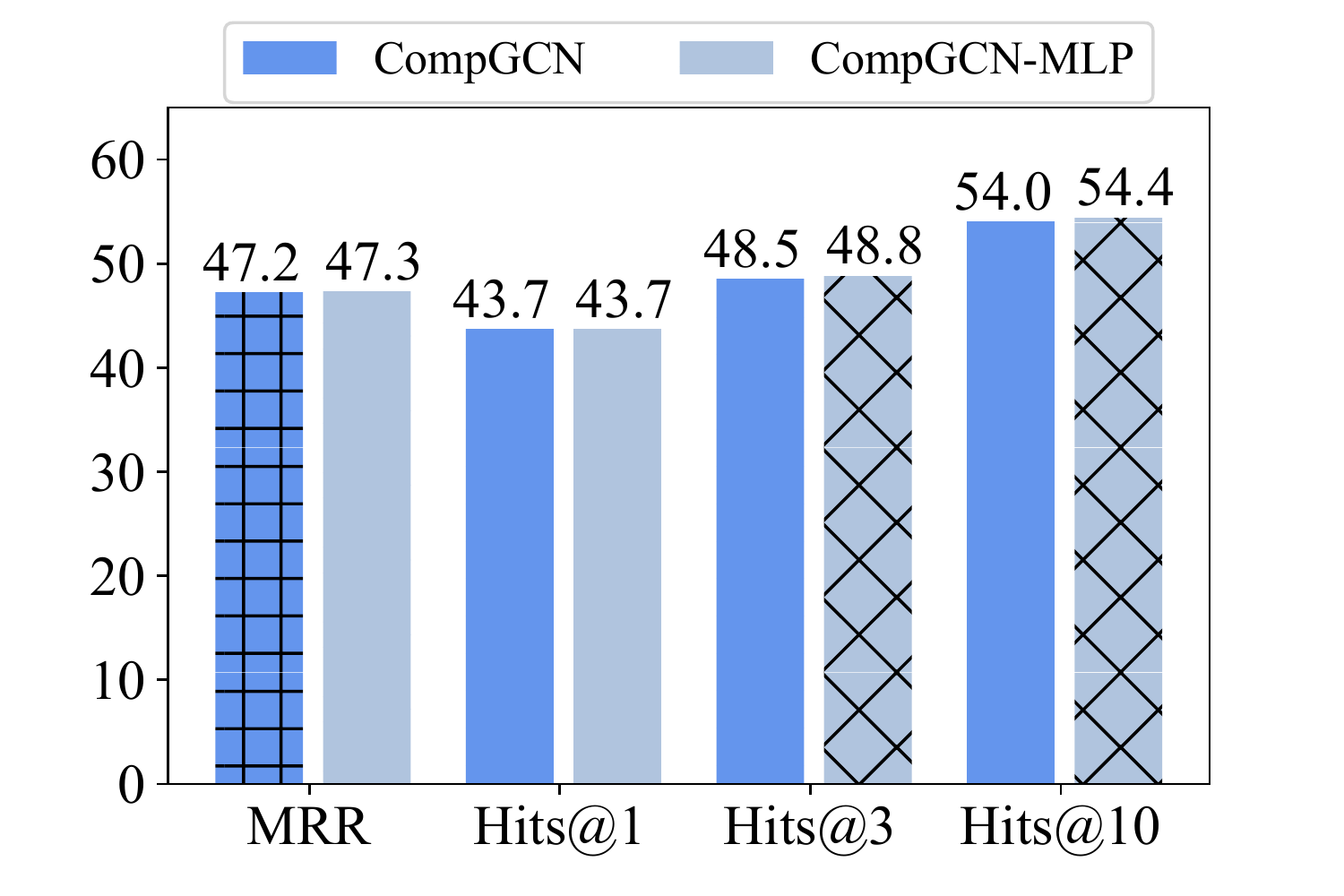}  }}
}
\vspace{-0.15in}
\caption{KGC results of CompGCN and CompGCN-MLP on \wnrr. On this dataset, CompGCN-MLP achieves compare performance as CompGCN.
}
\vspace{-0.2in}
\label{fig:com_finding_wn18rr}
\end{center}

\end{figure}

 \begin{figure}[t]
\begin{center}
 \centerline{

{\subfigure[\wn]
{\includegraphics[width=0.5\linewidth]{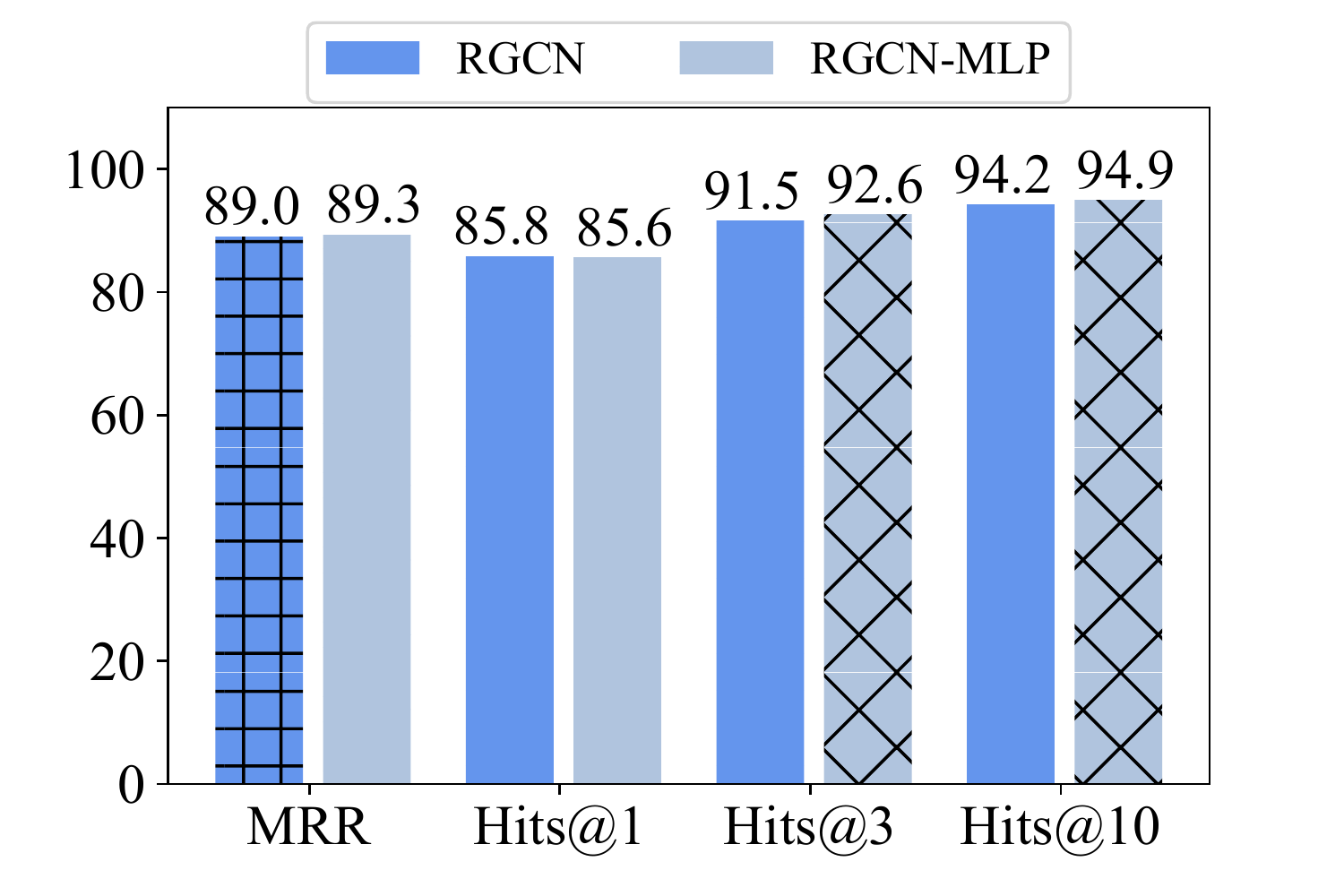}  }}

{\subfigure[\fb]
{\includegraphics[width=0.5\linewidth]{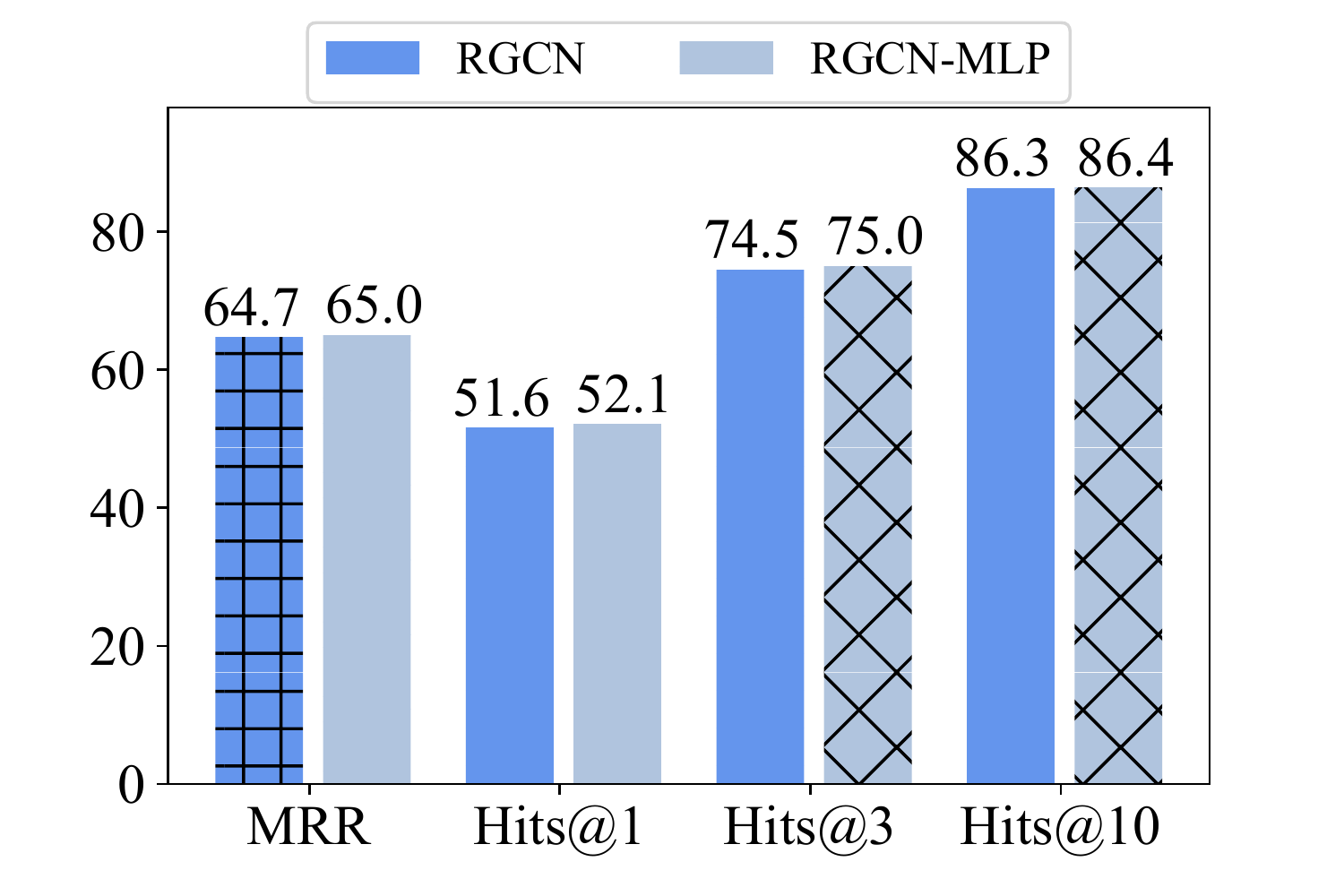}  }}
}
\vspace{-0.1in}
\caption{KGC results of RGCN and RGCN-MLP on \fb and \wn. On all three datasets, RGCN-MLP achieves comparable performance as RGCN. }
\vspace{-0.2in}
\label{fig:rgcn_finding_app}
\end{center}

\end{figure}

\begin{figure}[t]
\begin{center}
 \centerline{

{\subfigure[\wnrr]
{\includegraphics[width=0.5\linewidth]{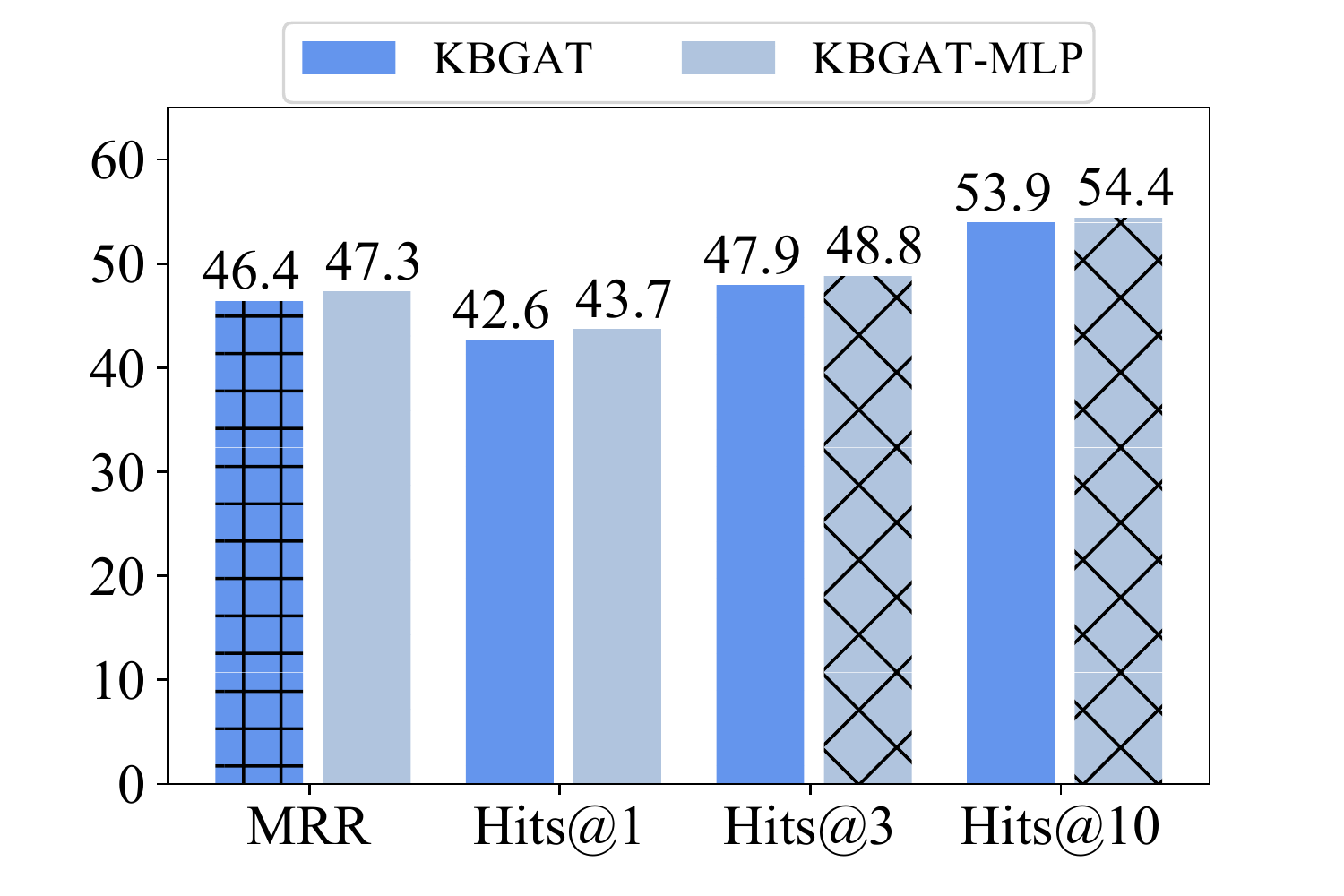}  }}

{\subfigure[\nell]
{\includegraphics[width=0.5\linewidth]{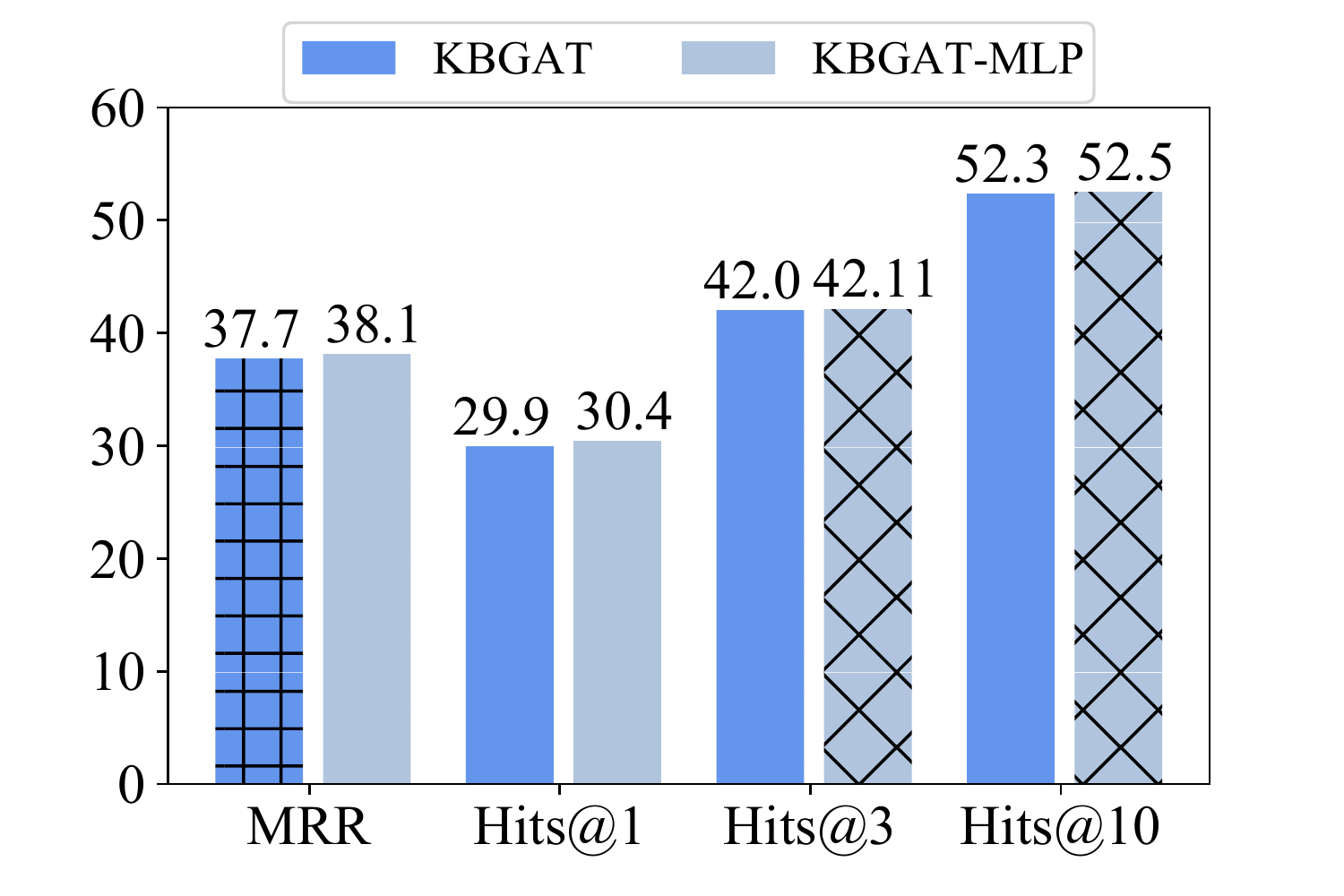}  }}
}
\vspace{-0.1in}
\caption{KGC results of KBGAT and KBGAT-MLP on   \wnrr and \nell. On all three datasets, KBGAT-MLP achieves comparable performance as KBGAT. }
\label{fig:kbgat_finding_app}
\end{center}

\end{figure}

In section~\ref{sec:finding},
We replace the message passing with the MLP while keeping other components untouched in CompGCN, RGCN and KBGAT. Due to the space limit, we only present the resutls on the \fbt dataset in section~\ref{sec:finding}. In this section, we include additional results on other datasets. Specifically, we include results of CompGCN/CompGCN-MLP on the \wnrr dataset, RGCN/RGCN-MLP on the \wn and \fb dataset, KBGAT/KBGAT-MLP on the \wnrr and \nell in \figurename~\ref{fig:com_finding_wn18rr}, \figurename~\ref{fig:rgcn_finding_app} and \figurename~\ref{fig:kbgat_finding_app} respectively.  All the counterpart MLP-based models achieve similar performance with the corresponding MPNN-based models, which show similar observations with the \fbt datasets in section~\ref{sec:finding}.

\section{MPNNs with random graph structure for message passing }
\label{app:exp_noise}

In this section, we 
investigate how the performance perform when we use random generated graph structure in the message passing process. The number of random edges is the same as the ones in the original graph.
 When training the model by optimizing the loss function, we still use the original  graph structure, i.e., $\mathcal{D}^{*}_{train}$ in Eq.~(\ref{Eq:loss}) is fixed in Appendix~\ref{app:loss}.   
Note that if the message passing has some contribution to the performance, aggregating the random edges should lead to the performance drop.

We present the results of  CompGCN, RGCN and KBGAT on various datasets in Table~\ref{tab:noise_in_aggregation}. 
We use ``Original", ``Random" to denote the performance with the original graph structure and    random edges respectively. From Table~\ref{tab:noise_in_aggregation}, we observe that using the noise edges achieves comparable performance, which further indicates that the message passing component is not the key part.  

\section{ Time Complexity}
\label{app:time}
We first define the sizes of weight matrices and embeddings of a single layer. We denote the dimension of entity and relation embeddings as $d$. The weight matrices in RGCN and CompGCN (shown in Eqs.~\eqref{eq:rgcn} and \eqref{eq:compgcn} respectively in Appendix~\ref{app:aggreation}. ) 
are $d\times d$ matrices. In KBGAT (Eq.~\eqref{eq:kbgat} in Appendix~\ref{app:aggreation}),  there are two weight matrices $\mathbf{W}_1$ and $\mathbf{W}_2$  of size $d\times 3d$ and $1 \times d$, respectively. Thus, the time complexity of RGCN, CompGCN, KBGAT for a single layer is $O(|E|d^2 +nd^2)$, $O(|E|d^2 +nd^2)$, $O(3|E|d^2 +nd^2+|E|d)$, respectively, where $|E|$ is the number of edges and $n$ is the number of nodes. While MLP doesn’t have the message passing operation, the time complexity in a single layer is $O(nd^2)$. Note $|E|$ is usually much larger than $n$, thus the MLP is more efficient than MPNN.

\section{Ensembling MLPs}
\label{app:ensemble}
We briefly introduce the MLP-based models we utilized for constructing the ensemble model in section~\ref{sec:ensemble} for the three datasets as follows:
\begin{compactitem}[\textbullet]
\item For the \fbt dataset, we ensemble the following models: DistMult + \emph{w/o sampling}; DistMult + \emph{with sampling} (two different settings with the number of negative samples as $0.5N$ and $N$, respectively); ConvE + \emph{w/o sampling}; ConvE + \emph{with sampling} (five different settings with the number of negative samples as $50, 200, 500,$ $0.5N$ and $N$, respectively). 
\item For the \wnrr dataset, we ensemble the following models: DistMult + \emph{w/o sampling}; ConvE+ \emph{w/o sampling}; ConvE+ \emph{with sampling} (one setting with the number of negative samples as $N$). 
\item For the \nell dataset, we ensemble the following models: ConvE+ \emph{w/o sampling};
ConvE+\emph{with sampling} (five settings with the number of negative samples as $50, 200, 500,0.5N$ and $N$).
\end{compactitem}

%% file: main.bbl
\begin{thebibliography}{34}
\expandafter\ifx\csname natexlab\endcsname\relax\def\natexlab#1{#1}\fi

\bibitem[{Bollacker et~al.(2008)Bollacker, Evans, Paritosh, Sturge, and
  Taylor}]{bollacker2008freebase}
Kurt Bollacker, Colin Evans, Praveen Paritosh, Tim Sturge, and Jamie Taylor.
  2008.
\newblock Freebase: a collaboratively created graph database for structuring
  human knowledge.
\newblock In \emph{Proceedings of the 2008 ACM SIGMOD international conference
  on Management of data}, pages 1247--1250.

\bibitem[{Bordes et~al.(2013)Bordes, Usunier, Garc{\'{\i}}a{-}Dur{\'{a}}n,
  Weston, and Yakhnenko}]{BordesUGWY13}
Antoine Bordes, Nicolas Usunier, Alberto Garc{\'{\i}}a{-}Dur{\'{a}}n, Jason
  Weston, and Oksana Yakhnenko. 2013.
\newblock Translating embeddings for modeling multi-relational data.
\newblock In \emph{Advances in Neural Information Processing Systems 26: 27th
  Annual Conference on Neural Information Processing Systems 2013. Proceedings
  of a meeting held December 5-8, 2013, Lake Tahoe, Nevada, United States},
  pages 2787--2795.

\bibitem[{Carlson et~al.(2010)Carlson, Betteridge, Kisiel, Settles, Hruschka,
  and Mitchell}]{carlson2010toward}
Andrew Carlson, Justin Betteridge, Bryan Kisiel, Burr Settles, Estevam~R
  Hruschka, and Tom~M Mitchell. 2010.
\newblock Toward an architecture for never-ending language learning.
\newblock In \emph{Twenty-Fourth AAAI conference on artificial intelligence}.

\bibitem[{Dettmers et~al.(2018)Dettmers, Minervini, Stenetorp, and
  Riedel}]{ettmersMS018}
Tim Dettmers, Pasquale Minervini, Pontus Stenetorp, and Sebastian Riedel. 2018.
\newblock Convolutional 2d knowledge graph embeddings.
\newblock In \emph{Proceedings of the Thirty-Second {AAAI} Conference on
  Artificial Intelligence, (AAAI-18), the 30th innovative Applications of
  Artificial Intelligence (IAAI-18), and the 8th {AAAI} Symposium on
  Educational Advances in Artificial Intelligence (EAAI-18), New Orleans,
  Louisiana, USA, February 2-7, 2018}, pages 1811--1818. {AAAI} Press.

\bibitem[{Fan et~al.(2019)Fan, Ma, Li, He, Zhao, Tang, and Yin}]{fan2019graph}
Wenqi Fan, Yao Ma, Qing Li, Yuan He, Eric Zhao, Jiliang Tang, and Dawei Yin.
  2019.
\newblock Graph neural networks for social recommendation.
\newblock In \emph{The World Wide Web Conference}, pages 417--426.

\bibitem[{Fellbaum(2010)}]{fellbaum2010wordnet}
Christiane Fellbaum. 2010.
\newblock Wordnet.
\newblock In \emph{Theory and applications of ontology: computer applications},
  pages 231--243. Springer.

\bibitem[{Garc{\'\i}a-Dur{\'a}n et~al.(2018)Garc{\'\i}a-Dur{\'a}n,
  Duman{\v{c}}i{\'c}, and Niepert}]{garcia2018learning}
Alberto Garc{\'\i}a-Dur{\'a}n, Sebastijan Duman{\v{c}}i{\'c}, and Mathias
  Niepert. 2018.
\newblock Learning sequence encoders for temporal knowledge graph completion.
\newblock \emph{arXiv preprint arXiv:1809.03202}.

\bibitem[{Jin et~al.(2021)Jin, Zhao, Zhang, Liu, Tang, and Shah}]{jin2021graph}
Wei Jin, Lingxiao Zhao, Shichang Zhang, Yozen Liu, Jiliang Tang, and Neil Shah.
  2021.
\newblock Graph condensation for graph neural networks.
\newblock \emph{arXiv preprint arXiv:2110.07580}.

\bibitem[{Jin et~al.(2019)Jin, Qu, Jin, and Ren}]{jin2019recurrent}
Woojeong Jin, Meng Qu, Xisen Jin, and Xiang Ren. 2019.
\newblock Recurrent event network: Autoregressive structure inference over
  temporal knowledge graphs.
\newblock \emph{arXiv preprint arXiv:1904.05530}.

\bibitem[{Lin et~al.(2015)Lin, Liu, Sun, Liu, and Zhu}]{lin2015learning}
Yankai Lin, Zhiyuan Liu, Maosong Sun, Yang Liu, and Xuan Zhu. 2015.
\newblock Learning entity and relation embeddings for knowledge graph
  completion.
\newblock In \emph{Twenty-ninth AAAI conference on artificial intelligence}.

\bibitem[{Liu et~al.(2021)Liu, Dai, So, and Le}]{liu2021pay}
Hanxiao Liu, Zihang Dai, David~R So, and Quoc~V Le. 2021.
\newblock Pay attention to mlps.
\newblock \emph{arXiv preprint arXiv:2105.08050}.

\bibitem[{Ma et~al.(2021)Ma, Liu, Zhao, Liu, Tang, and Shah}]{ma2021unified}
Yao Ma, Xiaorui Liu, Tong Zhao, Yozen Liu, Jiliang Tang, and Neil Shah. 2021.
\newblock A unified view on graph neural networks as graph signal denoising.
\newblock In \emph{Proceedings of the 30th ACM International Conference on
  Information \& Knowledge Management}, pages 1202--1211.

\bibitem[{Nathani et~al.(2019)Nathani, Chauhan, Sharma, and
  Kaul}]{nathani2019learning}
Deepak Nathani, Jatin Chauhan, Charu Sharma, and Manohar Kaul. 2019.
\newblock Learning attention-based embeddings for relation prediction in
  knowledge graphs.
\newblock In \emph{Proceedings of the 57th Annual Meeting of the Association
  for Computational Linguistics}, pages 4710--4723.

\bibitem[{Schlichtkrull et~al.(2018)Schlichtkrull, Kipf, Bloem, Van Den~Berg,
  Titov, and Welling}]{schlichtkrull2018modeling}
Michael Schlichtkrull, Thomas~N Kipf, Peter Bloem, Rianne Van Den~Berg, Ivan
  Titov, and Max Welling. 2018.
\newblock Modeling relational data with graph convolutional networks.
\newblock In \emph{European semantic web conference}, pages 593--607. Springer.

\bibitem[{Sun et~al.(2020)Sun, Vashishth, Sanyal, Talukdar, and
  Yang}]{sun2020re}
Zhiqing Sun, Shikhar Vashishth, Soumya Sanyal, Partha Talukdar, and Yiming
  Yang. 2020.
\newblock A re-evaluation of knowledge graph completion methods.
\newblock In \emph{Proceedings of the 58th Annual Meeting of the Association
  for Computational Linguistics}, pages 5516--5522.

\bibitem[{Tolstikhin et~al.(2021)Tolstikhin, Houlsby, Kolesnikov, Beyer, Zhai,
  Unterthiner, Yung, Keysers, Uszkoreit, Lucic et~al.}]{tolstikhin2021mlp}
Ilya Tolstikhin, Neil Houlsby, Alexander Kolesnikov, Lucas Beyer, Xiaohua Zhai,
  Thomas Unterthiner, Jessica Yung, Daniel Keysers, Jakob Uszkoreit, Mario
  Lucic, et~al. 2021.
\newblock Mlp-mixer: An all-mlp architecture for vision.
\newblock \emph{arXiv preprint arXiv:2105.01601}.

\bibitem[{Toutanova and Chen(2015)}]{toutanova2015observed}
Kristina Toutanova and Danqi Chen. 2015.
\newblock Observed versus latent features for knowledge base and text
  inference.
\newblock In \emph{Proceedings of the 3rd workshop on continuous vector space
  models and their compositionality}, pages 57--66.

\bibitem[{Toutanova et~al.(2015)Toutanova, Chen, Pantel, Poon, Choudhury, and
  Gamon}]{toutanova2015representing}
Kristina Toutanova, Danqi Chen, Patrick Pantel, Hoifung Poon, Pallavi
  Choudhury, and Michael Gamon. 2015.
\newblock Representing text for joint embedding of text and knowledge bases.
\newblock In \emph{Proceedings of the 2015 conference on empirical methods in
  natural language processing}, pages 1499--1509.

\bibitem[{Vashishth et~al.(2020)Vashishth, Sanyal, Nitin, and
  Talukdar}]{VashishthSNT20}
Shikhar Vashishth, Soumya Sanyal, Vikram Nitin, and Partha~P. Talukdar. 2020.
\newblock Composition-based multi-relational graph convolutional networks.
\newblock In \emph{8th International Conference on Learning Representations,
  {ICLR} 2020, Addis Ababa, Ethiopia, April 26-30, 2020}. OpenReview.net.

\bibitem[{Wang et~al.(2019)Wang, He, Cao, Liu, and Chua}]{wang2019kgat}
Xiang Wang, Xiangnan He, Yixin Cao, Meng Liu, and Tat-Seng Chua. 2019.
\newblock Kgat: Knowledge graph attention network for recommendation.
\newblock In \emph{Proceedings of the 25th ACM SIGKDD International Conference
  on Knowledge Discovery \& Data Mining}, pages 950--958.

\bibitem[{West et~al.(2014)West, Gabrilovich, Murphy, Sun, Gupta, and
  Lin}]{west2014knowledge}
Robert West, Evgeniy Gabrilovich, Kevin Murphy, Shaohua Sun, Rahul Gupta, and
  Dekang Lin. 2014.
\newblock Knowledge base completion via search-based question answering.
\newblock In \emph{Proceedings of the 23rd international conference on World
  wide web}, pages 515--526.

\bibitem[{Xiong et~al.(2017{\natexlab{a}})Xiong, Power, and
  Callan}]{xiong2017explicit}
Chenyan Xiong, Russell Power, and Jamie Callan. 2017{\natexlab{a}}.
\newblock Explicit semantic ranking for academic search via knowledge graph
  embedding.
\newblock In \emph{Proceedings of the 26th international conference on world
  wide web}, pages 1271--1279.

\bibitem[{Xiong et~al.(2017{\natexlab{b}})Xiong, Hoang, and Wang}]{XiongHW17}
Wenhan Xiong, Thien Hoang, and William~Yang Wang. 2017{\natexlab{b}}.
\newblock Deeppath: {A} reinforcement learning method for knowledge graph
  reasoning.
\newblock In \emph{Proceedings of the 2017 Conference on Empirical Methods in
  Natural Language Processing, {EMNLP} 2017, Copenhagen, Denmark, September
  9-11, 2017}, pages 564--573. Association for Computational Linguistics.

\bibitem[{Yang et~al.(2015)Yang, Yih, He, Gao, and Deng}]{YangYHGD14a}
Bishan Yang, Wen{-}tau Yih, Xiaodong He, Jianfeng Gao, and Li~Deng. 2015.
\newblock Embedding entities and relations for learning and inference in
  knowledge bases.
\newblock In \emph{3rd International Conference on Learning Representations,
  {ICLR} 2015, San Diego, CA, USA, May 7-9, 2015, Conference Track
  Proceedings}.

\bibitem[{Ye et~al.(2019)Ye, Li, Fang, Zang, and Wang}]{ye2019vectorized}
Rui Ye, Xin Li, Yujie Fang, Hongyu Zang, and Mingzhong Wang. 2019.
\newblock A vectorized relational graph convolutional network for
  multi-relational network alignment.
\newblock In \emph{IJCAI}, pages 4135--4141.

\bibitem[{Ying et~al.(2018)Ying, He, Chen, Eksombatchai, Hamilton, and
  Leskovec}]{ying2018graph}
Rex Ying, Ruining He, Kaifeng Chen, Pong Eksombatchai, William~L Hamilton, and
  Jure Leskovec. 2018.
\newblock Graph convolutional neural networks for web-scale recommender
  systems.
\newblock In \emph{Proceedings of the 24th ACM SIGKDD International Conference
  on Knowledge Discovery \& Data Mining}, pages 974--983.

\bibitem[{Yu et~al.(2021)Yu, Yang, Zhang, and Wu}]{yu2021knowledge}
Donghan Yu, Yiming Yang, Ruohong Zhang, and Yuexin Wu. 2021.
\newblock Knowledge embedding based graph convolutional network.
\newblock In \emph{Proceedings of the Web Conference 2021}, pages 1619--1628.

\bibitem[{Zhang et~al.(2022{\natexlab{a}})Zhang, Liu, Sun, and
  Shah}]{zhang2021graphless}
Shichang Zhang, Yozen Liu, Yizhou Sun, and Neil Shah. 2022{\natexlab{a}}.
\newblock Graph-less neural networks: Teaching old mlps new tricks via
  distillation.
\newblock \emph{ICLR}.

\bibitem[{Zhang and Yao(2022)}]{zhang2022knowledge}
Yongqi Zhang and Quanming Yao. 2022.
\newblock Knowledge graph reasoning with relational digraph.
\newblock In \emph{Proceedings of the ACM Web Conference 2022}, pages 912--924.

\bibitem[{Zhang et~al.(2022{\natexlab{b}})Zhang, Wang, Ye, and
  Wu}]{zhang2022rethinking}
Zhanqiu Zhang, Jie Wang, Jieping Ye, and Feng Wu. 2022{\natexlab{b}}.
\newblock Rethinking graph convolutional networks in knowledge graph
  completion.
\newblock In \emph{Proceedings of the ACM Web Conference 2022}, pages 798--807.

\bibitem[{Zhao et~al.(2022)Zhao, Jin, Akoglu, and Shah}]{zhao2021stars}
Lingxiao Zhao, Wei Jin, Leman Akoglu, and Neil Shah. 2022.
\newblock From stars to subgraphs: Uplifting any gnn with local structure
  awareness.
\newblock \emph{ICLR}.

\bibitem[{Zhu et~al.(2022)Zhu, Yuan, Xhonneux, Zhang, Gazeau, and
  Tang}]{zhu2022learning}
Zhaocheng Zhu, Xinyu Yuan, Louis-Pascal Xhonneux, Ming Zhang, Maxime Gazeau,
  and Jian Tang. 2022.
\newblock Learning to efficiently propagate for reasoning on knowledge graphs.
\newblock \emph{arXiv preprint arXiv:2206.04798}.

\bibitem[{Zhu et~al.(2021)Zhu, Zhang, Xhonneux, and Tang}]{zhu2021neural}
Zhaocheng Zhu, Zuobai Zhang, Louis-Pascal Xhonneux, and Jian Tang. 2021.
\newblock Neural bellman-ford networks: A general graph neural network
  framework for link prediction.
\newblock \emph{Advances in Neural Information Processing Systems}, 34.

\bibitem[{Zoph and Le(2016)}]{zoph2016neural}
Barret Zoph and Quoc~V Le. 2016.
\newblock Neural architecture search with reinforcement learning.
\newblock \emph{arXiv preprint arXiv:1611.01578}.

\end{thebibliography}
